\documentclass[acmtog]{acmart}
\acmSubmissionID{367}

\usepackage{booktabs} %
\usepackage{bm}
\usepackage[normalem]{ulem}
\usepackage{multirow}
\usepackage{caption}
\usepackage{subcaption}

\usepackage{wrapfig}

\definecolor{kuigreen}{rgb}{0.2, 0.8, 0.2}

\definecolor{darkgreen}{rgb}{0.2, 0.6, 0.2}
\newcommand{\revise}[1]{#1}
\newcommand{\reviserm}[1]{}

\usepackage[ruled]{algorithm2e} %

\usepackage{placeins} %

\SetAlFnt{\small}
\SetAlCapFnt{\small}
\SetAlCapNameFnt{\small}
\SetAlCapHSkip{0pt}

\setcopyright{rightsretained}
\acmJournal{TOG}
\acmYear{2021}\acmVolume{40}\acmNumber{4}\acmArticle{132}\acmMonth{8} \acmDOI{10.1145/3450626.3459832}

\begin{document}
\title{DiffAqua: A Differentiable Computational Design Pipeline for Soft Underwater Swimmers with Shape Interpolation}

\author{Pingchuan Ma}
\affiliation{%
 \institution{MIT CSAIL}
 \city{Cambridge}
 \state{MA}
 \country{USA}}
\email{pcma@csail.mit.edu}

\author{Tao Du}
\orcid{0000-0001-7337-7667}
\affiliation{%
 \institution{MIT CSAIL}
 \city{Cambridge}
 \state{MA}
 \country{USA}}
\email{taodu@csail.mit.edu}

\author{John Z. Zhang}
\affiliation{%
 \institution{ETH Zurich}
 \city{Zurich}
 \country{Switzerland}}
\email{john.zhang@srl.ethz.ch}

\author{Kui Wu}
\affiliation{%
 \institution{MIT CSAIL}
 \city{Cambridge}
 \state{MA}
 \country{USA}}
\email{kuiwu@csail.mit.edu}

\author{Andrew Spielberg}
\affiliation{%
 \institution{MIT CSAIL}
 \city{Cambridge}
 \state{MA}
 \country{USA}}
\email{aespielberg@csail.mit.edu}

\author{Robert K. Katzschmann}
\affiliation{%
 \institution{ETH Zurich}
 \city{Zurich}
 \country{Switzerland}}
\email{rkk@ethz.ch}

\author{Wojciech Matusik}
\affiliation{%
 \institution{MIT CSAIL}
 \city{Cambridge}
 \state{MA}
 \country{USA}}
\email{wojciech@csail.mit.edu}

\renewcommand\shortauthors{Ma et al.}

\begin{abstract}
The computational design of soft underwater swimmers is challenging because of the high degrees of freedom in soft-body modeling. In this paper, we present a differentiable pipeline for co-designing a soft swimmer's geometry and controller. Our pipeline unlocks gradient-based algorithms for discovering novel swimmer designs more efficiently than traditional gradient-free solutions. We propose Wasserstein barycenters as a basis for the geometric design of soft underwater swimmers since it is differentiable and can naturally interpolate between bio-inspired base shapes \emph{via} optimal transport. By combining this design space with differentiable simulation and control, we can efficiently optimize a soft underwater swimmer's performance with fewer simulations than baseline methods. We demonstrate the efficacy of our method on various design problems such as fast, stable, and energy-efficient swimming and demonstrate applicability to multi-objective design.

\end{abstract}



\begin{teaserfigure}
    \centering
    \includegraphics[width=\textwidth]{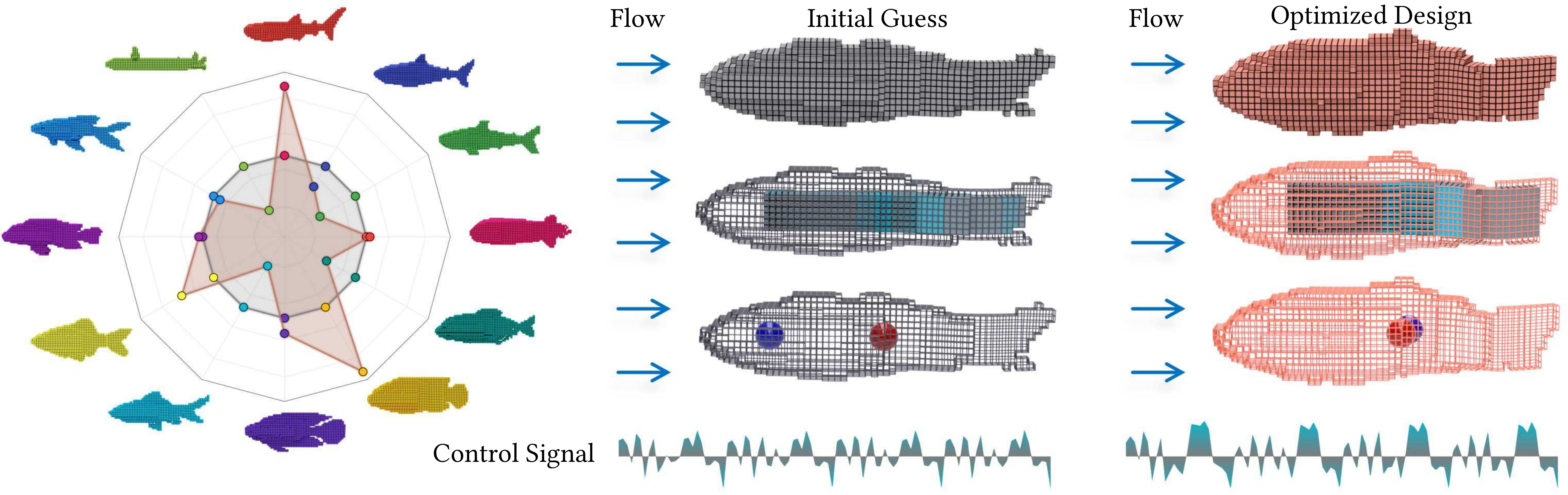}
    \vspace{-1.em}
    \caption{In this example, we co-design a swimmer's geometry and controller for position-keeping in running water (blue arrows in upper middle and upper right). The blue and red spheres indicate the desired and actual position of the swimmer. The swimmer's geometric design weights are initialized as the average of 12 base shapes with the Wasserstein distance metric shown in gray (left). After optimization, the final geometric design weights of each base shape are shown in orange (left). With the initial shape and actuator locations (upper middle) and the initial controller (lower middle), the position-keeping performance is poor, as indicated by the large distance between the red and blue dots. With the optimized shape and actuator locations (upper right) and the optimized controller (lower right), performance is markedly improved, \emph{i.e.}, blue and red are closer together.}
    \label{fig:teaser}
\end{teaserfigure}

\maketitle

\section{Introduction}

Designing bio-inspired underwater swimmers has long been an exciting interdisciplinary research problem for biologists and engineers~\cite{triantafyllou1995efficient,marchese2014autonomous,katzschmann2018exploration, berlinger2018modular, fish2006passive}. Aquatic locomotive performance of underwater swimmers is governed by two interrelated aspects: the control policy responsible for coordinated actuation and the geometry by which the actuation is transformed into motion through hydrodynamic forces. While the computational co-design of geometry and control has been explored in the context of articulated rigid walking and flying robots~\cite{zhao2020,du2016,pathak2019learning,ha2019reinforcement}, the co-design of geometry and control of a \emph{soft} robot consisting of substantially deformable materials has been sparsely studied. Since a soft robot is governed by continuum physics with many more degrees of freedom (DOFs), existing methods are not readily transferable to soft robotic design problems. Optimizing over all degrees of freedom of an infinite-dimensional continuum elastic body is computationally intractable, even when approximated with a large number of discrete elements. Lower-dimensional geometric representations are needed to allow for efficient shape exploration without sacrificing expressiveness.

To remedy the lack of an intuitive and low-dimensional design space suitable for soft underwater swimmers, we propose representing a soft swimmer's shape and actuators as probability distributions in 3D space and interpolate between designs with optimal transport~\cite{rubner2000earth}. Given a set of base shapes representing swimmer archetypes, we define a vector space spanned by these shapes according to the Wasserstein distance metric~\cite{solomon2015}. Such a representation of the soft swimmer's design space brings a few key benefits. First, it allows for each design to be represented as a low-dimensional Wasserstein barycentric weight vector. Second, and more importantly, this interpolation procedure is differentiable~\revise{\cite{bonneel2016wasserstein}}, enabling gradient-based optimization algorithms for fast exploration of the design space. We further combine this design space with a differentiable simulator~\cite{du2021diffpd} and control policy, creating a full pipeline for co-optimizing the body and brain of soft swimmers using gradient-based optimization algorithms.

We evaluate our co-optimization pipeline on a set of soft swimmer design problems whose objectives include forward swimming, stable swimming under opposing flows, and energy conservation. We further demonstrate the pipeline's applicability to multi-objective design and the generation of Pareto fronts. We show that our algorithm converges significantly faster than gradient-free optimization algorithms~\cite{hansen2003reducing} and strategies that alternate between optimizing geometric design and control.

In this paper, we contribute:
\begin{itemize}
    \item a low-dimensional differentiable design space parametrizing the shape and actuation of soft underwater swimmers with the Wasserstein distance metric,
    \item a differentiable pipeline for co-optimizing the geometric design and controller of a soft swimmer concurrently, and
    \item demonstrations of this algorithm on a set of bio-inspired single- and multi-objective underwater swimming tasks.
\end{itemize}

\section{Related Work} 

Our approach builds upon recent and seminal work in differentiable simulation, soft robot and character control, computational co-design of robots, and shape parametrizations.

\paragraph{Differentiable simulation}
Differentiable simulation allows for the direct computation of the gradients of continuous parameters affecting the system performance, such as control, material, and geometric parameters.
Gradients computed from differentiable simulation can be directly fed into numerical optimization algorithms, immediately unlocking applications such as system identification, computational control, and design optimization. Differentiable simulation has a rich history in robotics and physically-based animation across various domains, including
 rigid-body dynamics \cite{popovic2003motion, degrave2019differentiable, de2018end, geilinger2020add}, fluid dynamics \cite{holl2020learning, mcnamara2004fluid, schenck2018spnets} and cloth physics with rigid coupling \cite{liang2019differentiable, qiao2020scalable}. In cases where manually deriving gradients is difficult, automatic differentiation frameworks \cite{giftthaler2017automatic, hu2019difftaichi} or learned, approximate models \cite{li2018learning, sanchez2020learning, chen2018neural} have been employed. Most related to our work are differentiable simulation methods for soft bodies \cite{hu2019chainqueen, hahn2019real2sim, du2021diffpd, huang2021plasticinelab}. We build upon the work of \citet{du2021diffpd}, which proposes a differentiable projective dynamics framework capable of implicit integration.

\paragraph{Dynamic soft robot and character control}
Soft robotic control is traditionally more difficult than rigid robotic control, since the infinite-dimensional state spaces of soft robots are difficult to computationally reason about.
\revise{Previous papers \cite{geijtenbeek2013flexible, won2019learning} demonstrated learning general controllers for articulated, rigid-body walking robots in a variety of forms.} \citet{tan2011articulated} explored model-based control optimization of articulated rigid swimmers in an environment coupled with fluid. \revise{Control using model-free reinforcement learning for systems with solid-fluid coupling has been studied in \citet{ma2018fluid}}.
Simplified dynamical models which treat soft tendril structures compactly and accurately as rod-and-spring-like structures have allowed for the natural adoption of modern control algorithms like those used for articulated rigid robots \cite{marchese2016dynamics,della2018dynamic,katzschmann2019dynamic}.
\revise{\citet{grzeszczuk1995automated} proposed representing the actuators of swimmers with a simplified biomechanical model and automatically generating their controllers.}
\revise{There are also works, such as \citet{hecker2008real}, synthesizing animation independent of character morphology for the purpose of downstream retargeting, with the motion described by familiar
posing and key-framing methods.} More general approaches \cite{barbivc2008real,barbivc2009deformable,thieffry2018control,spielberg2019learning,katzschmann2019dynamically} present dimensionality reduction strategies for soft robots to compact representations used in model-based control. 
Most similar to our work are the approaches of \citet{hu2019chainqueen} and \citet{min2019softcon}.  In \citet{hu2019chainqueen}, a differentiable soft-body simulator is coupled with parametric neural network controllers.  By tracking the positions and velocities of manually specified regions as control inputs, a loss function measuring the  forward progress in robot locomotion is directly backpropagated to control parameters, enabling efficient model-based control optimization \emph{via} gradient descent. \citet{min2019softcon} present model-free control of soft swimmers using reinforcement learning with similar handcrafted features. Our approach for control optimization is similar to the model-based optimization framework of \citet{hu2019chainqueen}, but instead of considering legs and crawlers, we optimize underwater swimmers. Our work further distinguishes itself from prior art through the realization of the non-trivial co-optimization of shape and control of soft underwater swimmers.

\paragraph{Computational robot co-design}

Much of the existing work on the co-design of rigid robots has focused on the interplay between geometric, inertial and control parameters.  \citet{megaro2015interactive, schulz2017interactive} explored the interactive design of robot control and geometry, in which CAD-like front-ends guided human-in-the-loop, simulation-driven control and design optimization.
\citet{wampler2009optimal, spielberg2017functional, ha2017joint} presented algorithms for co-optimizing geometric and inertial parameters of robots with open-loop controllers; \citet{schaff2019jointly, ha2019reinforcement} extended these ideas to the space of closed-loop neural network controllers \emph{via} reinforcement learning approaches. \citet{du2016} presented a method for co-optimizing the control and geometric parameters for optimal multicopter performance. \revise{\citet{sims1994evolving} models the morphology as a graph and co-optimizes it with control using an evolutionary strategy.} In each of the approaches described above, the  geometric representations are simple. Each robotic link is parameterized by no more than a handful of geometric parameters, substantially limiting the morphological search.  
More complex representations are used in the works by \citet{ha2018computational, wang2019neural, zhao2020}, presenting algorithms for searching over various robot topologies.  These techniques lead to more geometrically and functionally varied robots;  however, this discrete search space cannot be directly optimized by the continuous optimization approaches to which differentiable simulation lends itself.  

Computational co-design of soft robots has been more sparsely explored.  \citet{cheney2014unshackling, corucci2016evolving, van2019spatial} presented heuristic search algorithms for searching over soft robotic topology and control, including swimmers in the case of \citet{corucci2016evolving}.  These sampling-based approaches focus on geometry and (discrete) materiality, typically leaving actuation as open-loop, pre-programmed cyclic patterns.  Contrasted with these approaches are those of \citet{hu2019chainqueen, spielberg2019learning}, which exploit system gradients to co-optimize over closed-loop control, observation models, and spatially varying material parameters, but not geometry.  
Our approach combines advantages of both lines of research, relying on the differentiable simulation for fast gradient-based co-optimization of neural network control and complex geometry.  %

\paragraph{Shape interpolation}
Many bases for shape representations have been proposed over the years for applications in shape analysis.
Geometrically-based spaces \cite{ovsjanikov2012functional, lewis2014practice, baek2015isometric, solomon2015, schulz2017retrieval, bonneel2016wasserstein} parameterize shape collections based on intrinsic geometric metrics computed across a data set; learning-based spaces \cite{bronstein2011shape, averkiou2014shapesynth, fish2014meta, yang2019pointflow, mo2019structurenet, park2019deepsdf}, by contrast, learn parametrizations based on statistical features of that dataset and can also easily incorporate non-geometric information (such as labels).  In this work, we opt for the geometrically-based convolutional Wasserstein basis \cite{solomon2015}, which interpolates between meshes using the Wasserstein distance.  This basis smoothly interpolates between shapes with ``as few'' in-between modifications as necessary, keeping our mesh parametrizations well-behaved and minimizing the chance of artifacts that might cause difficulties in simulation.  This continuous and differentiable basis makes it amenable to continuous co-optimization.  %

\section{System Overview}

\begin{figure*}[ht]
  \centering
  \includegraphics[width=\linewidth]{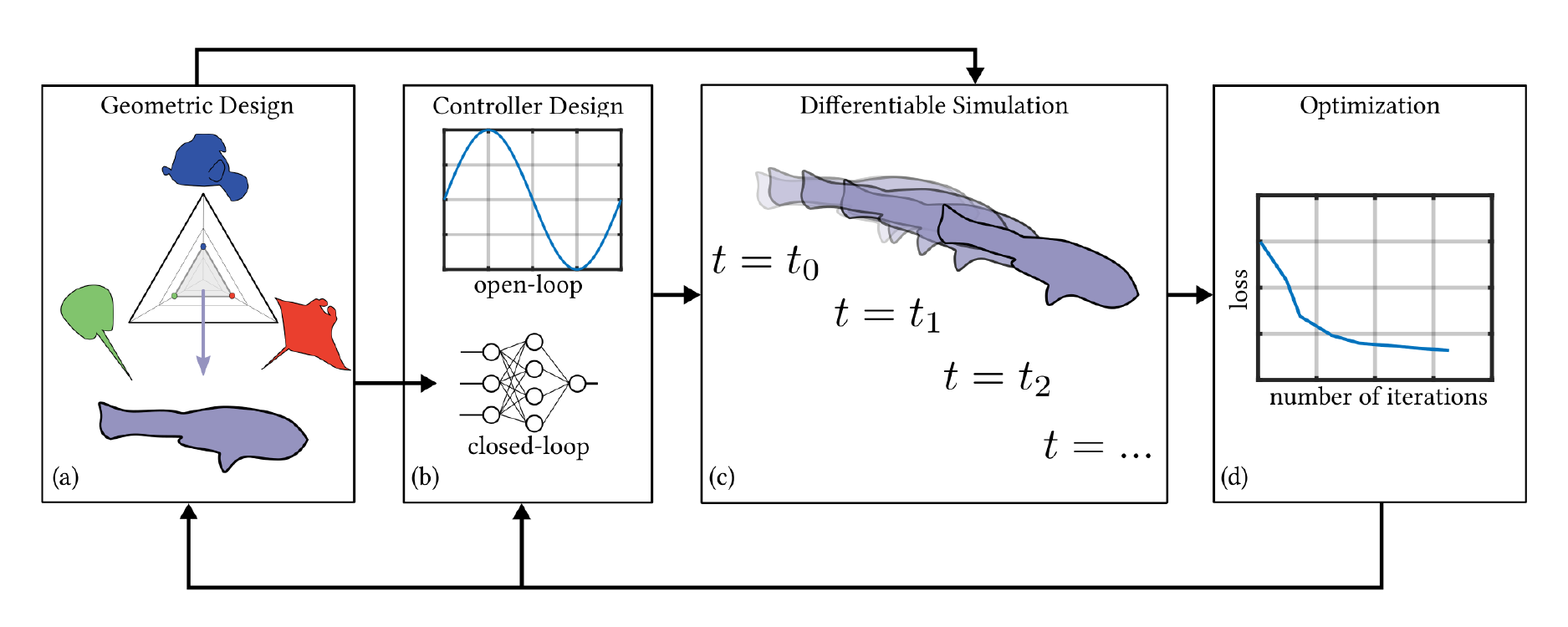}
  \vspace{-2.em}
  \caption{In our computational design pipeline, we begin with (a) the shape and actuator design using the Wasserstein distance to interpolate smoothly between base shapes and between actuators. Given a set of base shapes by the user, we initialize the shape parametrization by assigning equal weights to all bases. (b) Next, we consider the controller design for two cases: an open-loop controller and a closed-loop controller. (c) Given this initial shape, actuators, and controller, we simulate the design using a differentiable simulator to evaluate its performance. (d) Finally, we take advantage of the differentiability of our framework to compute the gradients of a given objective with respect to geometry and control parameters simultaneously. These gradients are used in a gradient descent optimizer to concurrently adapt the geometric and control design parameters.}
  \label{fig:overview}
\end{figure*}

Our design optimization procedure starts with a collection of base shapes from which to build a geometric design space using the Wasserstein distance metric.  Each base design also specifies a multivariate normal distribution for each of its actuators. For our bio-inspired exploration of swimmers, we have selected base shapes inspired by nature (Sec.~\ref{sec:design}), \emph{e.g.}, sharks, manta rays, and goldfish. Next, the user specifies a controller which is either an open-loop sinusoidal signal or a neural network. The state observations of the closed-loop neural network controller are defined by the user (Sec.~\ref{sec:control}). The user then evaluates the soft swimmer \emph{via} differentiable simulation (Sec.~\ref{sec:simulation}), which returns both the swimmer's performance and its gradients with respect to geometric design and control parameters. Finally, we embed both the design space and the differentiable simulator in a gradient-based optimization procedure, which co-optimizes both the geometric design and the controller until convergence (Sec.~\ref{sec:optimization}), closing our design loop. We present the overview of our method in Fig.~\ref{fig:overview}.

\section{Geometric Design}\label{sec:design}

Our soft swimmer's geometric design consists of two parts: its volumetric shape and its actuator locations. This section explains our differentiable parameterization of this geometric design space. The key idea is to represent geometric designs as probability density functions and generate novel swimmers by interpolating the probability density functions of the base designs. After the design space has been parametrized, any swimmer in this space can be represented compactly by low-dimensional discrete and continuous parameters.

\subsection{Shape Design}
Given a set of user-defined base shapes, we interpolate between these bases with the Wasserstein distance metric to form a continuous and differentiable shape space. The reason behind this choice is twofold. First, the Wasserstein barycentric interpolation encourages smooth, plausible intermediate results even between shapes with substantially different topology~\cite{solomon2015}, which is common in underwater creatures. Second, efficient numerical solutions exist for solving Wasserstein barycenters, and their gradients are also readily available~\revise{\cite{solomon2015,bonneel2016wasserstein}}. These two features make Wasserstein distance a proper choice for building a fully differentiable shape design space for soft swimmers.

Formally, we consider the designs of our soft swimmers to be embedded in an axis-aligned bounding box $\Omega\subseteq\mathcal{R}^d$, where $d=2$ or $3$ is the dimension. Without loss of generality, we rescale $\Omega$ uniformly and shift it so that it has unit volume and is centered at the origin of $\mathcal{R}^d$. We use $d_E: \Omega\times\Omega\rightarrow\mathcal{R}_+$ to denote the Euclidean distance function, $\mathcal{P}(\Omega)$ the space of probability measures on $\Omega$, and $\mathcal{P}(\Omega\times\Omega)$ the space of probability measures on $\Omega\times\Omega$.

For any probability measure $P\in\mathcal{P}(\Omega)$, we define the following set $S\subseteq\Omega$ based on $P$'s probability density function $p$ to represent a soft swimmer's shape:
\begin{align}
S(p) = \{\bm{x}|p(\bm{x})\geq0.5 \sup_{\bm{x}\in\Omega} p(\bm{x})\}.
\end{align}
In other words, a soft swimmer's shape occupies the volumetric region where the probability density is over half the peak density. Similarly, the surface of the shape, which is primarily used for computing hydrodynamic forces and visualizing the shape, is defined as follows:
\begin{align}
\partial S(p) = \{\bm{x}|p(\bm{x})=0.5 \sup_{\bm{x}\in\Omega} p(\bm{x})\}
\end{align}

\paragraph{Shape bases} Although these probability density functions provide us a design space that can express almost all possible soft swimmer shape designs, its infinite degrees of freedom makes design optimization computationally challenging. Moreover, the probability measure space includes many physically implausible shapes that would be rejected instantly by a human user. To encourage findings of physically plausible swimmers from a low-dimensional shape space, we build a library consisting of $m$ soft swimmers designed by human experts and define the shape space as the vector space spanned from these designs. Specifically, let $S_1,S_2,\cdots,S_m\subseteq\Omega$ be the $m$ shapes in the library, we define the probability density function $p_i$ for $i=1,2,\cdots,m$ as follows:
\begin{align}
p_i(\bm{x})=\begin{cases}
\frac{1}{|S_i|} & \text{if }\bm{x}\in S_i, \\
0 & \text{otherwise.}
\end{cases}
\end{align}
Furthermore, we use $P_i$ to refer to the corresponding probability measures induced by $p_i$, which serve as the basis for shape interpolation.

\paragraph{Shape interpolation} With all probability measures $P_i$ at hand, we consider the shape space parametrized by a weight vector $\bm{\alpha}\in\mathcal{R}^m$ in the probability simplex $\{\bm{\alpha}|\alpha_i\geq0,\sum_i \alpha_i=1\}$. Specifically, given a weight vector $\bm{\alpha}$, we compute the Wasserstein barycenter $P(\bm{\alpha})$, which can be interpreted as a weighted average of $P_1,P_2,\cdots,P_m$ \cite{solomon2015}:
\begin{align}
P(\bm{\alpha})=\underset{P\in\mathcal{P}(\Omega)}{\arg\min}\sum_i \alpha_i \mathcal{W}_2^2(P,P_i).
\end{align}
Here, $\mathcal{W}_2(\cdot,\cdot):\mathcal{P}(\Omega)\times\mathcal{P}(\Omega)\rightarrow\mathcal{R}_+$ is the 2-Wasserstein distance:
\begin{align}
\mathcal{W}_2(P,Q)=\left[\inf_{\pi\in\Pi(P,Q)} \iint_{\Omega\times\Omega} d_E^2(\bm{x},\bm{y})d\pi(\bm{x},\bm{y})\right]^{\frac{1}{2}}
\end{align}
where $\Pi(P,Q)$ is the set of transportation maps from probability measure P to Q:
\begin{align}
\Pi(P,Q)=\{\pi\in\mathcal{P}(\Omega\times\Omega)|\pi(\cdot,\Omega)=P,\pi(\Omega,\cdot)=Q\}.
\end{align}
In short, for each weight vector $\bm{\alpha}$, we compute the Wasserstein barycenter $P(\bm{\alpha})$ and use its probability density function to define a new shape $S(\bm{\alpha})$, with some abuse of notation $S$. This way, we establish an isomorphic mapping from the probability simplex to the shape space.   Exploring the shape space is then equivalent to navigating the continuous, low-dimensional probability simplex.

\subsection{Actuator Design}
In this work, we use the contractile muscle fiber model introduced by~\citet{min2019softcon} to actuate soft swimmers. Like the shape interpolation scheme above, users first specify the geometric representation of actuators for all soft swimmers in the library. Then, our method interpolates between them to obtain actuators for shapes at Wasserstein barycenters.

\begin{figure}[tp]
    \centering
    \includegraphics[width=\linewidth]{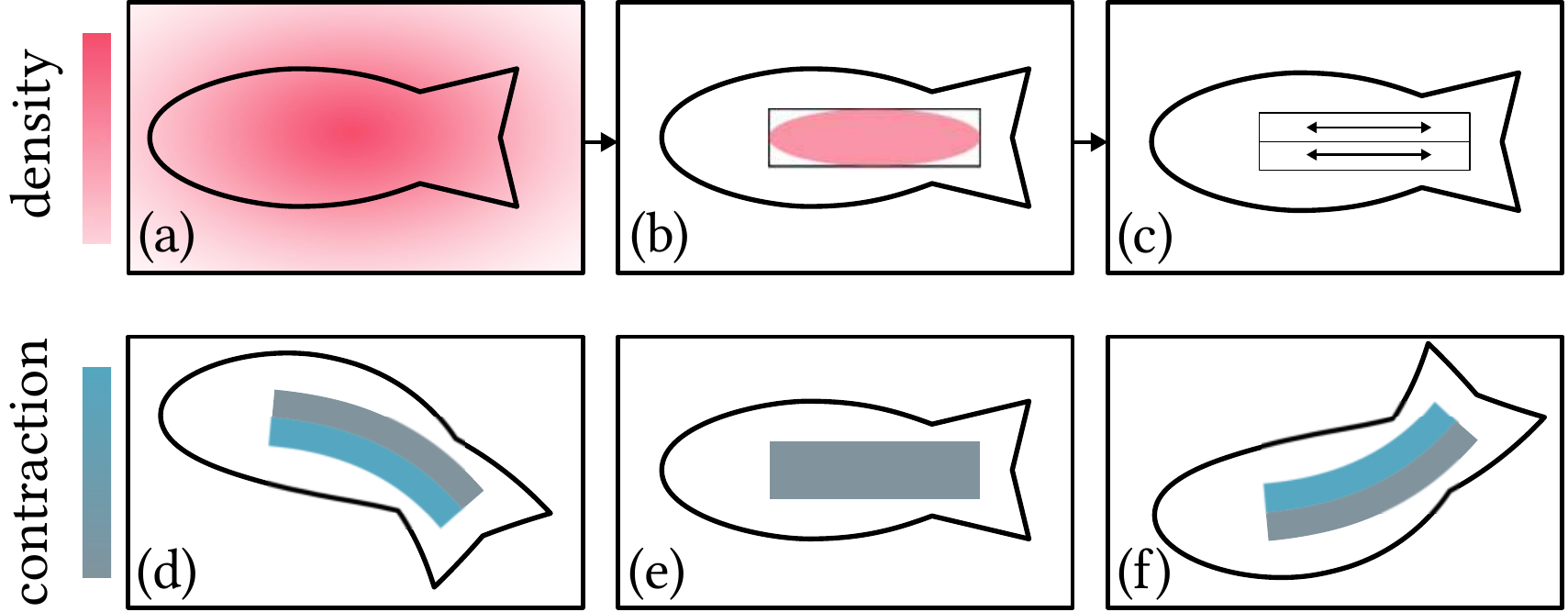}
    \caption{We explain our actuator design with a 2D fish illustration. Sec.~\ref{sec:design}: (a) We use a multivariate normal distribution to parametrize the location of the actuators. The gradient of light pink to dark pink represents the increasing probability density function. (b) The actuator's shape is defined by the bounding box of the probability density's isocontour at half of its peak density. We draw a black bounding box around the clipped distribution to form the actuator. Sec.~\ref{sec:control}: (c) Our actuator is divided into a pair of antagonistic muscle fibers acting in opposition. We show the extreme deformations of the fish tail in (d) the left-most position, (e) the neutral position, and (f) the right-most position. More contraction of each muscle fiber is indicated by a darker blue. The rest shape of the muscle fiber is depicted in gray.}
    \label{fig:muscle}
\end{figure}

\paragraph{Actuator representation} For a given soft swimmer, we define its actuators by a set of discrete and continuous labels. Each actuator has one discrete parameter denoting its category and a small number of continuous parameters defining its location, size, and orientation. The actuator categories supported in this work include ``left fin'', ``right fin'', and ``caudal fin'', which indicate the actuator's rough location (Fig.~\ref{fig:exp1}). Each soft swimmer has at most one actuator for each category. The actuator's continuous parameters represent its geometric design with a multivariate normal distribution $\mathcal{N}(\bm{\mu}, \bm{\Sigma})$ where $\bm{\mu}\in\mathcal{R}^d$ and $\bm{\Sigma}\in\mathcal{R}^{d\times d}$ are its mean and variance. Similar to the shape function $S$, the actuator's shape $A$ is defined by the locations whose probability density is over half of the peak density from $\mathcal{N}(\bm{\mu},\bm{\Sigma})$:
\begin{align}
A^c_S(\bm{\mu},\bm{\Sigma})=B(\{\bm{x}| \exp(-\frac{1}{2}(\bm{x}-\bm{\mu})^\top\bm{\Sigma}^{-1}(\bm{x}-\bm{\mu}))\geq0.5 \}) \cap S.
\end{align}
Here, $A^c_S$ stands for the volumetric region occupied by an actuator of category $c$ in a soft swimmer whose shape is $S$, and $B:\Omega\rightarrow\Omega$ is a function that takes as input a $d$-dimensional ellipsoid and returns the minimum bounding box whose directions are aligned with the principal axes of the ellipsoid (Fig.~\ref{fig:muscle}). Note that the intersection with $S$ ensures the actuator stays in the interior of $S$. With some abuse of notation, we use $A^c_P$ and $A^c_{\bm{\alpha}}$ to denote an actuator of category $c$ in a soft swimmer defined by a probability measure $P$ or the Wasserstein barycentric interpolation with a weight vector $\bm{\alpha}$, respectively.  When designing a soft swimmer $S_i$ in the library, the human expert also specifies its actuators' discrete and continuous parameters to obtain $\{A^c_i\}_c$, defined as follows:
\begin{align}
A^c_i=A^c_{S_i}(\bm{\mu}^c_i,\bm{\Sigma}^c_i)
\end{align}
where $\bm{\mu}^c_i$ and $\bm{\Sigma}^c_i$ are prespecified mean and variance for the actuators with category $c$ in shape $S_i$.

\paragraph{Actuator interpolation} We now describe how to generate actuators for a soft swimmer interpolated using the Wasserstein distance. Let $\bm{\alpha}$ be the weight vector defined above. For the soft swimmer defined by the probability measure $P(\bm{\alpha})$ and for each actuator category $c$, we define $A^c_{\bm{\alpha}}$ as follows:
\begin{align}
A^c_{\bm{\alpha}}=A^c_{S(\bm{\alpha})}(\bm{\mu}^c_{\bm{\alpha}},\bm{\Sigma}^c_{\bm{\alpha}})
\end{align}
where $\bm{\mu}^c_{\bm{\alpha}}$ and $\bm{\Sigma}^c_{\bm{\alpha}}$ are continuous parameters computed as follows:
\begin{align}
    \bm{\mu}^c_{\bm{\alpha}}=&~\text{LinearInterpolation}(\bm{\alpha}, \bm{\mu}^c_i), \\
    \bm{\Sigma}^c_{\bm{\alpha}}=&~(\bm{R}^c)^\top\bm{S}^c\bm{R}^c,\\
    \bm{R}^c=&~\text{RotationalInterpolation}(\bm{\alpha},\bm{R}^c_i),\\
    \bm{S}^c=&~\text{LinearInterpolation}(\bm{\alpha},\bm{S}^c_i).
\end{align}
Here, both $\bm{\mu}^c_{\bm{\alpha}}$ and $\bm{\Sigma}^c_{\bm{\alpha}}$ are defined by interpolating the means and variances of the actuators from the base shapes with the weight vector $\bm{\alpha}$. If the actuator of category $c$ is not used by the $i$-th base shape, we set the corresponding weight $\alpha_i$ to zero so that it is excluded from the interpolation. The mean $\bm{\mu}^c_{\bm{\alpha}}$ is computed by linearly interpolating between the \reviserm{$\bm{u}^c_i$} \revise{$\bm{\mu}^c_i$}. The variance matrix $\bm{\Sigma}^c_{\bm{\alpha}}$ is obtained by linearly interpolating the eigenvalues $\bm{S}^c_i$ and the Euler angles of the rotational matrices $\bm{R}^c_i$ from the actuator of category $c$ in the $i$-th base shapes. \revise{Note that, for simplicity, the actuators are not interpolated using the Wasserstein distance.  We leave solving for the exact actuator interpolation as future work.}

\section{Control}\label{sec:control}

In this section, we first explain the actuation model built upon the actuator's geometric design defined in the previous section. Next, we introduce two controllers for the soft underwater swimmer's motion: an open-loop controller defined by analytic functions and a closed-loop neural network controller.

\subsection{Actuation Model}

As we explain in the previous section, each actuator is a cubic region defined by a clipped multivariate normal distribution. Following the muscle fiber model from previous work~\citep{min2019softcon, du2021diffpd}, we model each actuator as a group of parallel muscle fibers. Each muscle fiber can contract itself along the fiber direction based on the magnitude of the control signal. For the ``caudal fin'' actuator, the parallel muscle fibers are placed antagonistically, allowing us to actuate bilateral flapping (Fig.~\ref{fig:muscle}).

\subsection{Open-Loop Controller}
The muscles of marine animals are usually actuated periodically like waves. %
Inspired by this, the first controller we consider in this work is a series of sinusoidal waves:
\begin{align}
a(t)=a\sin(\omega t + \varphi)
\end{align}
where $t$ stands for the time and $a$, $\omega$, and $\varphi$ are the control parameters to be optimized and may vary between different actuator categories.  When combined with the parallel muscle fibers in each actuator, the open-loop controller generates an oscillating motion sequence from the soft swimmer's body.

\subsection{Closed-Loop Controller}

In addition to open-loop controllers, we also consider using closed-loop neural network controllers to achieve more precise control over a soft swimmer's motion. Our neural network controller takes sensor data as input and returns control signals used to activate the actuation of the parallel muscle fibers.

\paragraph{Sensing} For a soft underwater swimmer in our design space, we gather position and velocity information from a swimmer's head, center, and tail. More concretely, we first align all swimmers in the library so that their heading is along the positive $x$-axis. After this alignment, we ensure the heading of any interpolated design is also along the positive $x$-axis. We then place the head, center, and tail sensor at locations along the $x$-axis with the maximum, zero, and minimum $x$ values within the swimmer's shape. We stack the outputs from all three sensors into a single vector and send it to the neural network controller.

\paragraph{Neural network controller} Our neural network controller is a standard multilayer perceptron network with two layers of $64$ neurons. We use $\tanh$ as the activation function. The input to the network includes the velocities from all three sensors and also the positional offsets from the center sensor to the head and tail sensors. Additionally, our network also takes as input a $20$-dimensional temporal encoding vector $\phi(t)$ to sense the temporally contextual information and encourage periodic control output, which is defined as follows~\cite{vaswani2017attention}:
\begin{align}
\begin{split}
\phi(t) = [&\sin(2^0\pi\tau(t)), \sin(2^1\pi\tau(t)), \cdots, \sin(2^9\pi\tau(t)), \\
&\cos(2^0\pi\tau(t)), \cos(2^1\pi\tau(t)), \cdots, \cos(2^9\pi\tau(t))].
\end{split}
\end{align}
Here, $\tau:\mathcal{R}_+\rightarrow[0,1]$ wraps the actual $t$ with a predefined period $T$: $\tau(t)=\frac{t\text{ mod }T}{T}$. We use $T=25h$ where $h$ is the time step in each experiment. \revise{We concatenate the sensor feedback and the temporal encoding as a $35$-dimensional vector, which is used as the input to the neural network controller.} The neural network then outputs control signals for all actuators.

\section{Differentiable Simulation}\label{sec:simulation}

Given a soft swimmer's geometric design and controller, we now describe how to evaluate the swimmer's performance and gradients in a differentiable simulation environment. We build our differentiable simulator upon~\citet{min2019softcon} and~\citet{du2021diffpd}, which use projective dynamics, a fast finite element simulation method amenable to implicit integration. We consider the geometric domain $\Omega$ as the material space and the swimmer's shape $S(p)$ as the rest shape of a deformable body. One way to forward simulate the design is to extract the surface boundary $\partial S(p)$ from the level set of its probability density function $p$, discretize $S(p)$ into finite elements, and simulate its motion by tracking its vertex locations. While this is a viable option for the forward simulation, this Lagrangian representation of the geometric design brings a few challenges for design optimization. First, whenever $S$ is updated, we need to regenerate the volumetric mesh to avoid narrow finite elements, and such a geometric processing step could be computationally expensive. Second, gradient computation depends on the exact discretization of $S(p)$, but the way to partition $S(p)$ into finite elements is not unique. Therefore, picking any specific partition may bias the gradient computation unintentionally.

\begin{figure}[t]
    \centering
    \includegraphics[width=0.9\linewidth]{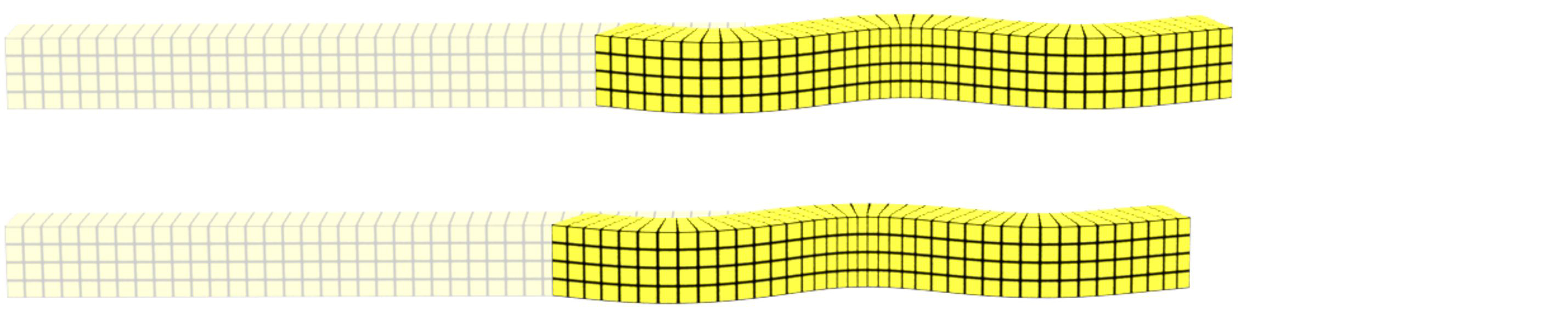}\vspace{1em}\\
    \includegraphics[width=\linewidth]{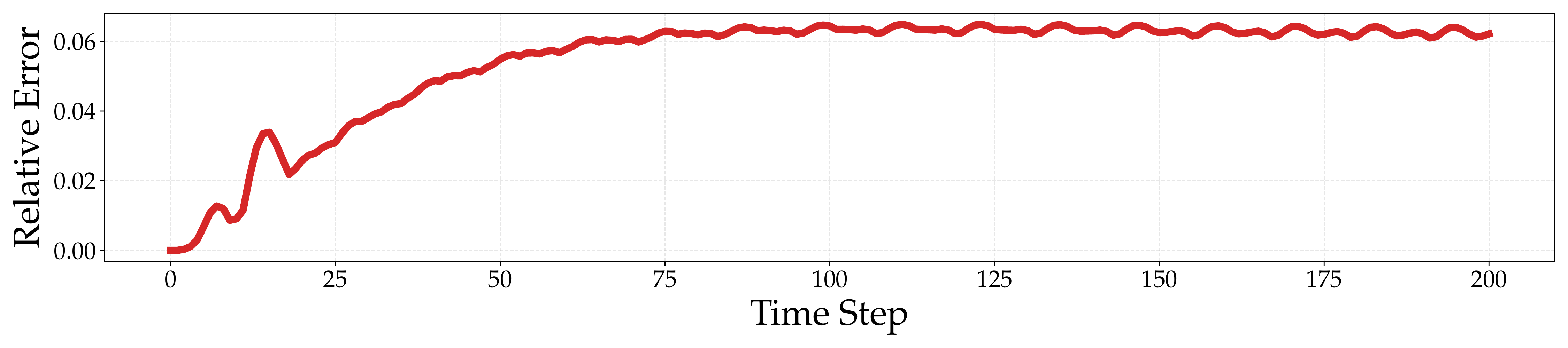}
    \vspace{-13em}
    \begin{flushright}{\small No exterior cells}\end{flushright}\vspace{1.5em}
    \begin{flushright}{\small Ours}\end{flushright}
    \vspace{6.5em}
    \caption{To understand the effects of simulating a volumetric region outside a swimmer with a close-to-zero Young's modulus, we simulate an eel without the volumetric region outside its shape (top) and compare it to our method (middle), which simulates the whole domain $\Omega$ but assigns small Young's modulus to the volumetric region outside the eel. The relative error for each time step converges to around $6\%$ after $100$ time steps (bottom). We compute the relative error by computing the average difference between nodal positions from the two simulations and dividing it by the length of the eel.}
    \label{fig:diff}
\end{figure}

\paragraph{Spatial and time discretization} As a result, we choose to evolve the soft swimmer's motion with an Eulerian view based on the probability density function $p$. Specifically, we simulate the full geometric domain $\Omega$ with a spatially varying stiffness defined by $p$. For the volumetric region outside $S(p)$, we assign close-to-zero stiffness so that simulating it has a negligible effect on the soft swimmer's motion. More formally, we consider a uniform grid that discretizes $\Omega$. Let $n$ be the number of nodes in the grid and let $\bm{q}_i,\bm{v}_i\in\mathcal{R}^{dn}$ be the nodal positions and velocities at the beginning of the $i$-th time step. We simulate the grid based on the implicit time-stepping scheme:
\begin{align}
\bm{q}_{i+1}=& \, \bm{q}_i +h\bm{v}_{i+1} \\
\bm{v}_{i+1}=& \, \bm{v}_i+h\bm{M}^{-1}[\bm{f}_{\text{int}}(\bm{q}_{i+1}) + \bm{f}_{\text{ext}}]
\end{align}
where $h$ denotes the time step, $\bm{M}\in\mathcal{R}^{dn\times dn}$ is the mass matrix,  and $\bm{f}_{\text{int}},\bm{f}_{\text{ext}}\in\mathcal{R}^{dn}$ are the internal and external forces applied to the $n$ nodes. As we inherit the constitutive model and the actuator model from~\citet{min2019softcon}, we skip their implementation details and focus on explaining how our spatially varying stiffness field is used to define the constitutive model. Specifically, we define a cell-wise constant Young's modulus field $E(c)$:
\begin{align}
E(c) = E_0 G(\hat{p}_c, 0.1)
\end{align}
where $c$ is the cell index, $E_0$ is a base Young's modulus and $\hat{p}_c$ is the discretized value of $p$ in cell $c$. We further divide $\hat{p}_c$ by its maximum value from all cells so that $\hat{p}_c$ is between $0$ and $1$. $G(\cdot,0.1):[0,1]\rightarrow[0,1]$ is the Schlick’s function~\cite{schlick1994fast}, \setlength{\intextsep}{0pt}%
\setlength{\columnsep}{10pt}%
\begin{wrapfigure}{r}{0.3\linewidth}%
    \centering
    \includegraphics[width=\linewidth]{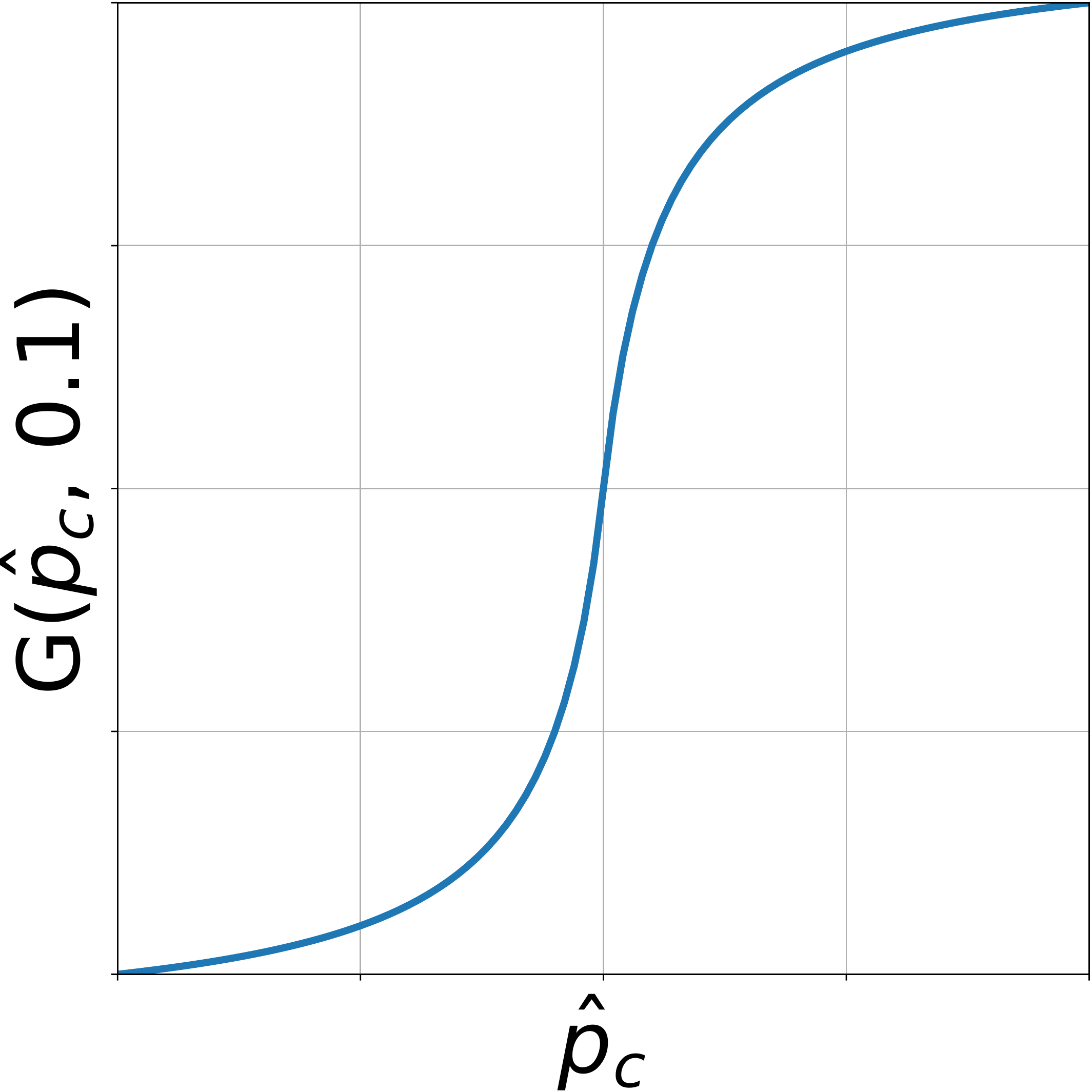}
\end{wrapfigure} and choosing its second argument to be $0.1$ pushes the output of $G$ towards the binary value $0$ or $1$ (see the inset). Recall that the soft swimmer's shape is defined at locations where $p$ is at least half of the peak density, using $G(\cdot, 0.1)$ encourages the soft swimmer's body to have a Young's modulus close to $E_0$. Additionally, it suppresses the volumetric region outside the swimmer to have a close-to-zero Young's modulus so that its effect on the swimmer's motion is negligible (Fig.~\ref{fig:diff}).

We stress that the choice of simulating the entire domain $\Omega$ with spatially varying Young's modulus brings us two key benefits. First, discretization is done trivially on the background grid. We avoid the expensive discretization process involving the level set of $p$, and evolving the shape design is merely updating the probability density function on the regular grid. Second, and more importantly, since a differentiable simulator provides gradients about the material stiffness, and algorithms for Wasserstein barycentric interpolation offer gradients about the probability density function~\revise{\cite{bonneel2016wasserstein}}, formulating the material stiffness as a function of the probability density connects the gradients from shape design and simulation into a uniform pipeline seamlessly.

\revise{
\paragraph{Hydrodynamics}
To efficiently mimic the interaction between the water and the swimmer, we develop a hydrodynamics formulation based on~\citet{min2019softcon}.
In this model, the thrust and drag forces are computed on each quadrilateral from the swimmer's surface mesh after discretization:
\begin{align}
    \bm{f}_{\text{drag}} &= \frac{1}{2} \rho A C_d (\Phi) \left\lVert \bm{v}_\text{rel} \right\rVert^2 \bm{d} , \label{eq:fdrag}\\
    \bm{f}_{\text{thrust}} &= -\frac{1}{2} \rho A C_t (\Phi) \left\lVert \bm{v}_\text{lat} \right\rVert^2 \bm{n} , \label{eq:fthrust}
\end{align}
where $A$ is the area of the surface quadrilateral, $\rho$ is the density of the fluid, $\bm{d}=\frac{\bm{v}_\text{rel}}{\lVert \bm{v}_\text{rel} \rVert}$ is the direction of the relative surface velocity, and $\bm{n}$ is the surface normal. $C_d(\Phi)$ and $C_t(\Phi)$ are dimensionless drag and thrust coefficients that only depend on the angle of attack $\Phi=\cos^{-1}(\bm{n}\cdot\bm{v}_\text{rel}) - \frac{\pi}{2}$. The relative velocity for the surface quadrilateral is calculated as follows:
\begin{equation}
    \bm{v}_\text{rel} = \bm{v}_\text{water} - \frac{1}{4} ( \bm{v}_0 + \bm{v}_1 + \bm{v}_2 + \bm{v}_3) ,
\end{equation}
where $\bm{v}_\text{water}$ is the velocity of the surrounding water and $\bm{v}_0$ to $\bm{v}_3$ are the velocities of the four corners of the quadrilateral. We define the lateral velocity in Eqn. (\ref{eq:fthrust}) as follows:
\begin{equation}
    \bm{v}_\text{lat} = \bm{v}_\text{rel} - (\bm{s}\cdot \bm{v}_\text{rel})\bm{s} ,
\end{equation}
where $\bm{s}$ is the direction of the fish spine. In our framework, $\bm{s}$ is set as an unit vector pointing from fish tail to the head. The thrust and drag forces ares then distributed equally to the four corners of each quadrilateral.}

\revise{
Lastly, the thrust force model from \citet{min2019softcon} creates an additional ``drag-like'' force even when there is no tail motion, which prevents forward swimming.  To alleviate this, we modify Eqn..~(\ref{eq:fthrust}), and scale the thrust by $\left\lVert \bm{v}_\text{lat} \right\rVert^2$, a physically-based modification which is inspired by the large amplitude elongated-body theory of fish lomocotion~\cite{lighthill1971large}.
}

\paragraph{Backpropagation} For a given soft swimmer's design and controller, a loss function provides a quantitative metric for evaluating the soft swimmer's performance. In this work, we define our loss function $L$ on nodal positions, velocities, and hydrodynamic forces. The choices of these input arguments allow us to measure travel distance, speed, or energy efficiency. We provide the definitions of each loss function in our experiments in Sec.~\ref{sec:results}.

After we evaluate the loss function for a given design and controller, we can compute their gradients \emph{via} backpropagation. To calculate the gradients about the design parameter $\bm{\alpha}$, we first obtain gradients about $E(c)$, the material stiffness inside each cell, from the differentiable simulator. We then backpropagate these gradients through Wasserstein barycentric interpolation to obtain $\frac{\partial L}{\partial\bm{\alpha}}$ through the mapping function $G(\cdot,0.1)$ as described above\revise{.
In particular, $\frac{\partial L}{\partial\bm{\alpha}}$ is computed using auto-differentiation as implemented by deep learning frameworks~\cite{paszke2019pytorch}}.  All of our controllers are also clearly differentiable, allowing gradients with respect to control parameters to also be computed \emph{via} backpropagation.

\section{Optimization}\label{sec:optimization}

Our fully differentiable pipeline enables the usage of gradient-based numerical optimization algorithms in co-designing the shape, actuator shape, and controller of soft underwater swimmers. Starting with an initial guess of the soft swimmer's geometric design and controller, we apply the Adam optimizer~\cite{kingma2014adam} with gradients computed for both design and control parameters. We terminate the optimization process when the results converge or when we exhaust a specified computational budget.  Since we parametrize both geometry and control with continuous variables in the same framework, 
we can optimize over all variables simultaneously. %

\section{Results}\label{sec:results}

\begin{table*}[htb]
    \centering
    \caption{\revise{A summary of all experiments mentioned in Sec.~\ref{sec:results}. The ``\# Bases'' and ``\# Parameters'' columns report the number of base shapes used for interpolation and the number of trainable parameters, respectively. The ``Resolution'' column gives the voxel resolution of base bounding boxes. The ``Shape'' and ``Control'' columns indicate whether we optimize a soft swimmer's geometric design and controller, respectively. The ``Objective'' column shows the experiment's objective type (none, single, or multiple). The ``Time'' column records the total time cost of optimization in minutes. %
    }}
    \label{tab:summary}
    \revise{
    \begin{tabular}{c|c|ccccccc}
    \toprule
    Section & Name & \# Bases & \# Parameters & Resolution & Shape & Control & Objective & Time (min) \\
    \midrule
    \multirow{2}{*}{8.1} & Shape Exploration & 3 & - & $60\times54\times14$ & - & - & None & - \\
    & Shape Optimization & 2 & 2 & $60\times16\times26$ & $\checkmark$ & - & Single & 136 \\
    \midrule
    \multirow{3}{*}{8.2} & Open-Loop Co-Optimization & 2 & 5 & $40\times8\times8$ & $\checkmark$ & $\checkmark$ & Single & \hspace{0.4em}60 \\ %
    & Closed-Loop Co-Optimization & 3 & 7572 & $60\times54\times14$ & $\checkmark$ & $\checkmark$ & Single & 354 \\
    & Large Dataset Co-Optimization & 12 & 6541 & $60\times16\times26$ & $\checkmark$ & $\checkmark$ & Single & 193 \\
    \midrule
    8.3 & Multi-Objective Co-Optimization & 12 & 6541 & $60\times16\times26$ & $\checkmark$ & $\checkmark$ & Multiple & 195 \\
    \bottomrule
    \end{tabular}
    }
\end{table*}

\begin{figure*}[htp]
    \centering
    \includegraphics[width=\linewidth]{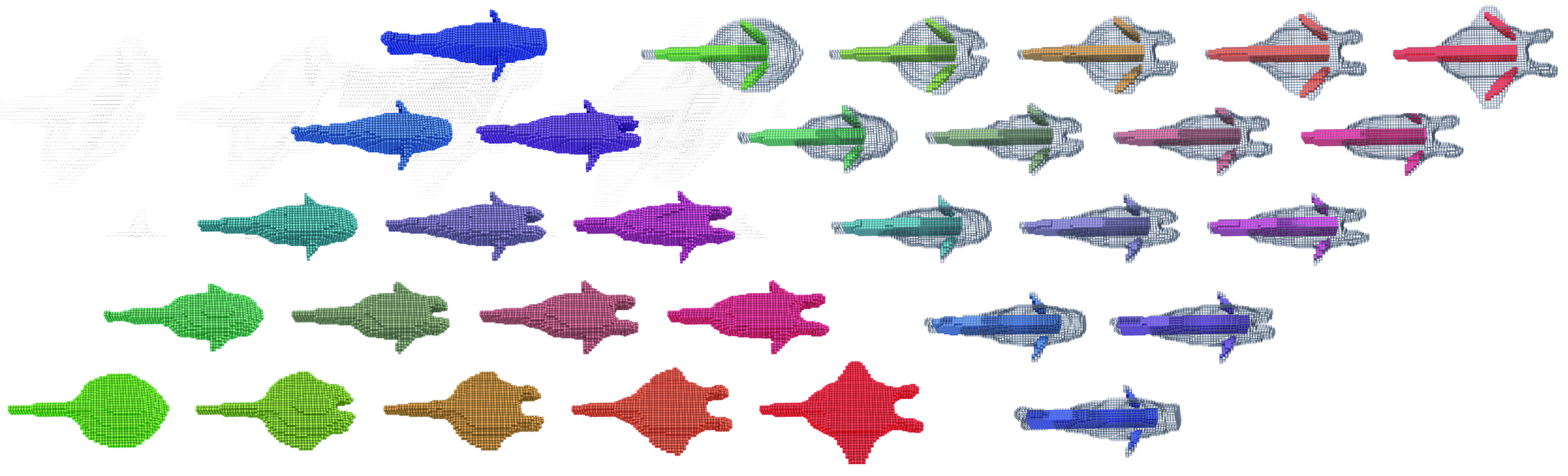}
    \caption{"Shape exploration" experiment: Interpolation between three base shapes and their actuators. Left triangle: 12 intermediate shapes interpolated from three base shapes (red, green, and blue shapes at the corners). Right triangle: the corresponding actuator placements (colored regions) for the 12 intermediate shapes shown as wire frames, also generated by interpolating actuators in the three base shapes at the corners.}
    \label{fig:exp1}
\end{figure*}

We evaluate our method's performance with six experiments, covering shape design, co-design of geometry and control, and multi-objective design. We summarize their setup in Table~\ref{tab:summary}. We run all experiments on a virtual machine instance from Google Cloud Platform with 16 Intel Xeon Scalable Processors (Cascade Lake) @ 3.1 GHz and 64 GB memory with 8 OpenMP threads in parallel.

\revise{We collect a set of meshes representing various fishes in nature as the design bases. In the pre-processing phase, we normalize and voxelize them using the same grid resolution.
Grid resolutions can be found in Table~\ref{tab:summary}.
W use the following strategies in all our experiments to improve the numerical stability of Wasserstein barycentric interpolation: choosing basis shapes with the same topological genus, aligning their centers of mass, and running the interpolation iteration until convergence.}

\subsection{Design Space Exploration}

\paragraph{Shape exploration} In this experiment, we demonstrate the capability of our shape interpolation scheme without considering optimization. In Fig.~\ref{fig:exp1}, we show shape and actuator interpolation between three morphologically different hand-designed base shapes from our library: a clownfish, a manta ray, and a stingray. We see from Fig.~\ref{fig:exp1} that the Wasserstein barycentric interpolation is capable of generating smooth and biologically plausible intermediate shapes. Additionally, we note that our actuator interpolation generates designs adaptive to the shape's size, as can be seen from the change of muscle size in the left and right fins.

\paragraph{Shape optimization} In this experiment, we demonstrate the power of our shape interpolation scheme when used in tandem with our differentiable simulator for optimizing shape design and actuator placement. We choose a shark with a tall tail and an orca with a flat tail as the base shapes. The control input is a prespecified open-loop sine wave leading to oscillatory motions on the horizontal plane (spanned by the $x$- and $y$-axes in our coordinate system), which is ill-suited for the orca but natural for the shark. We initialize the open-loop controller with $a=1$, $\omega=\frac{\pi}{6h}$, and $\varphi=0$. The objective is to find a geometric design that traverses the longest forward distance in a fixed time period $Nh$, where $N$ is the number of time steps and $h$ the time interval. We purposefully choose the shark and orca bases since we know \emph{a priori} that the shark will outperform the orca in this task with the prespecified controller. Therefore, this experiment also serves as a smell test for our pipeline. Formally, the loss function is defined as
\begin{align}\label{eq:loss_dist}
L=&-\frac{1}{|Sp|}\sum_{j\in Spine} (\bm{q}_{N,j})_x - (\bm{q}_{0,j})_x\\
Spine=&\{j|(\bm{q}_{0,j})_y=0\}
\end{align}
where $\bm{q}_{i,j}\in\mathcal{R}^d$ is the $j$-th node's location at the $i$-th time step and $(\cdot)_x$ and $(\cdot)_y$ extracts its $x$ and $y$ coordinate, respectively. \reviserm{The heading of all our swimmers' shape designs is along the positive $x$ axis. Therefore, maximizing its $x$-offset between the first and last time step, or minimizing its negation as in $L$, encourages a design that travels the longest forward distance.} \revise{The objective encourages faster swimming speed along the desired heading direction, which is defined as the $x$-axis in all our experiments.} We estimate the swimmer's $x$-offset by averaging the $x$-offsets between the first and last time step from nodes near the spine of the swimmer, which is formally defined by a $Spine$ set consisting of nodes whose $y$ coordinate (the lateral offset) is zero when undeformed.

\begin{figure}[htp]
    \centering
    \vspace{1em}
    \includegraphics[width=\linewidth]{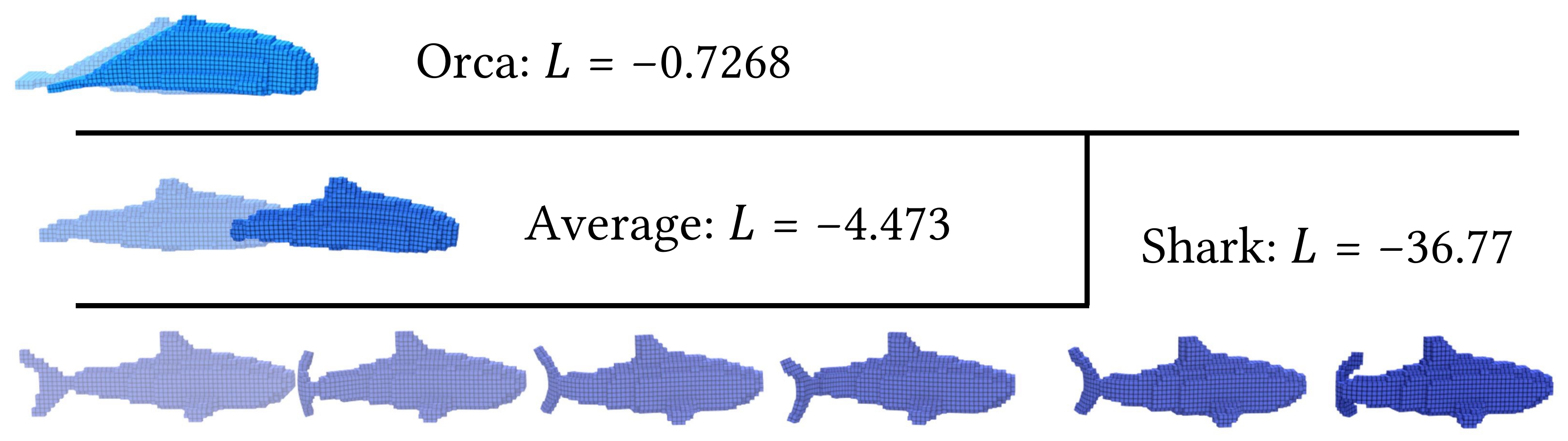}
    \caption{``Shape optimization'' experiment: Performances of two initial shapes and the optimized shape. Top: the orca; Middle: the average shape between the orca and the shark; Bottom: the optimized shape found by our method, which is similar to the shark. The color on the shape indicates the weight on the two base shapes (light blue: orca, dark blue: shark). A smaller loss $L$ indicates a longer traveling distance within the same duration and is preferred in this experiment.}
    \label{fig:exp2}
\end{figure}

Fig.~\ref{fig:exp2} summarizes our results in this experiment. We run our optimization algorithm with two different initial weight vectors $\bm{\alpha}$: $\bm{\alpha}=(0,1)$ which puts all weights in the orca (Fig.~\ref{fig:exp2} top) and $\bm{\alpha}=(0.5,0.5)$ (Fig.~\ref{fig:exp2} middle). As the control signal is known to be ill-suited for the orca, we expect the orca to have a poor performance, as is correctly reflected in the top row of Fig.~\ref{fig:exp2}. The optimized results from both starting shapes align with our expectation of the shark as the optimal shape for this task (Fig.~\ref{fig:exp2} bottom). Our algorithm robustly converges to the optimal solution in both cases.

\subsection{Co-Design of Shape and Control}

\begin{figure*}[htp]
    \centering
    \includegraphics[trim=100 0 0 100,clip,width=\linewidth]{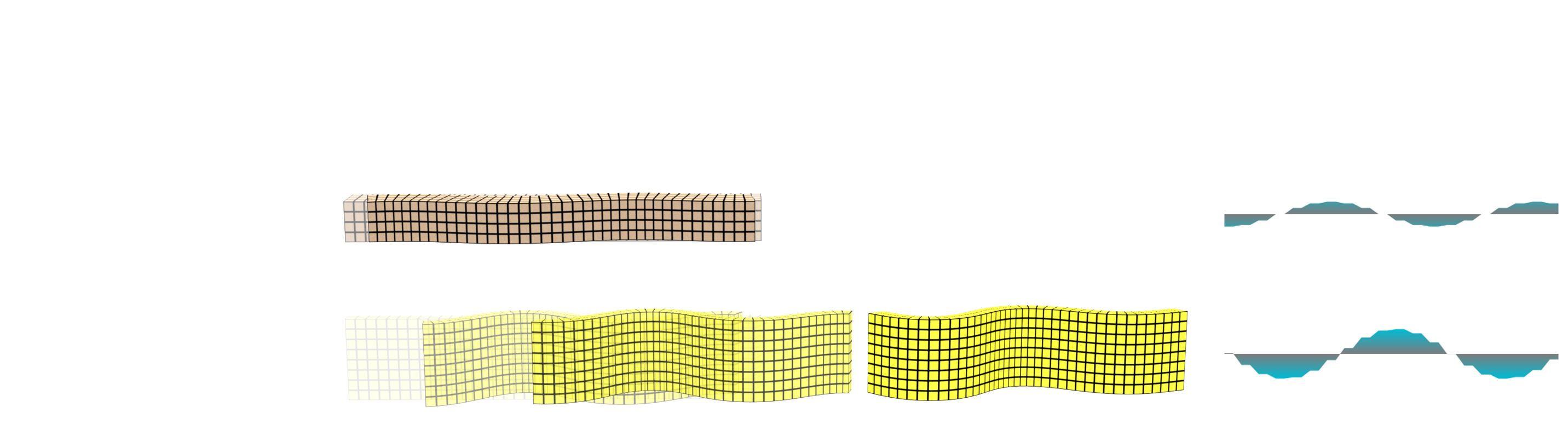}
    \vspace{-7.5em}
    \begin{flushleft}{\small \hspace*{0.25\linewidth}Shape\hspace*{0.5\linewidth}Control}\end{flushleft}\vspace{-4.5em}
    \begin{flushleft}{\small Init guess}\end{flushleft}\vspace{4.5em}
    \begin{flushleft}{\small Optimized}\end{flushleft}
    \vspace{1em}
    \caption{``Open-loop co-optimization'' experiment: The initial guess (top) and the optimized design (bottom) of the shape and control. The shapes are simulated to swim forward with the parameterized sinusoidal controller (right) from its initial position (transparent) to the final location (solid) within a fixed time period.}
    \label{fig:exp3}
\end{figure*}

\paragraph{Open-loop co-optimization} We now present our first co-design example, illustrating the value of our method compared to other baseline methods. The case we consider is co-optimization of both the shape and controller of an eel for the same objective described in the ``shape optimization'' experiment. The base shapes are two eels, \emph{i.e.}, slender bodies with their undulations' wavelengths smaller than their length. One \emph{vertical} eel is flag-like with a greater height than width, and the other \emph{horizontal} eel is pancake-like with a greater width than height. We use the same open-loop sine wave control sequence as in the ``shape optimization'' experiment, except that we leave its amplitude, phase, and frequency as variables to be optimized from the initial guess with $0.5$, $0$, and $\frac{\pi}{6h}$ respectively. The decision variables for this co-optimization problem are four-dimensional, including one geometric design parameter and three control parameters. The goal is to find both an optimal shape \textit{and} an optimal controller that leads to the longest distance traveled within a fixed time span. Fig.~\ref{fig:exp3} shows the initial design (top row) and the optimized design (bottom row) returned by our algorithm. Since the actuation  is manifested in form of a sine-wave on the horizontal plane, the optimal shape should be a vertical eel. As expected, our co-optimization algorithm correctly finds such a physical design, \revise{as well as an intensified control signal to maximize traveling distance}.

\begin{figure}[!htp]
    \centering
    \includegraphics[width=\linewidth]{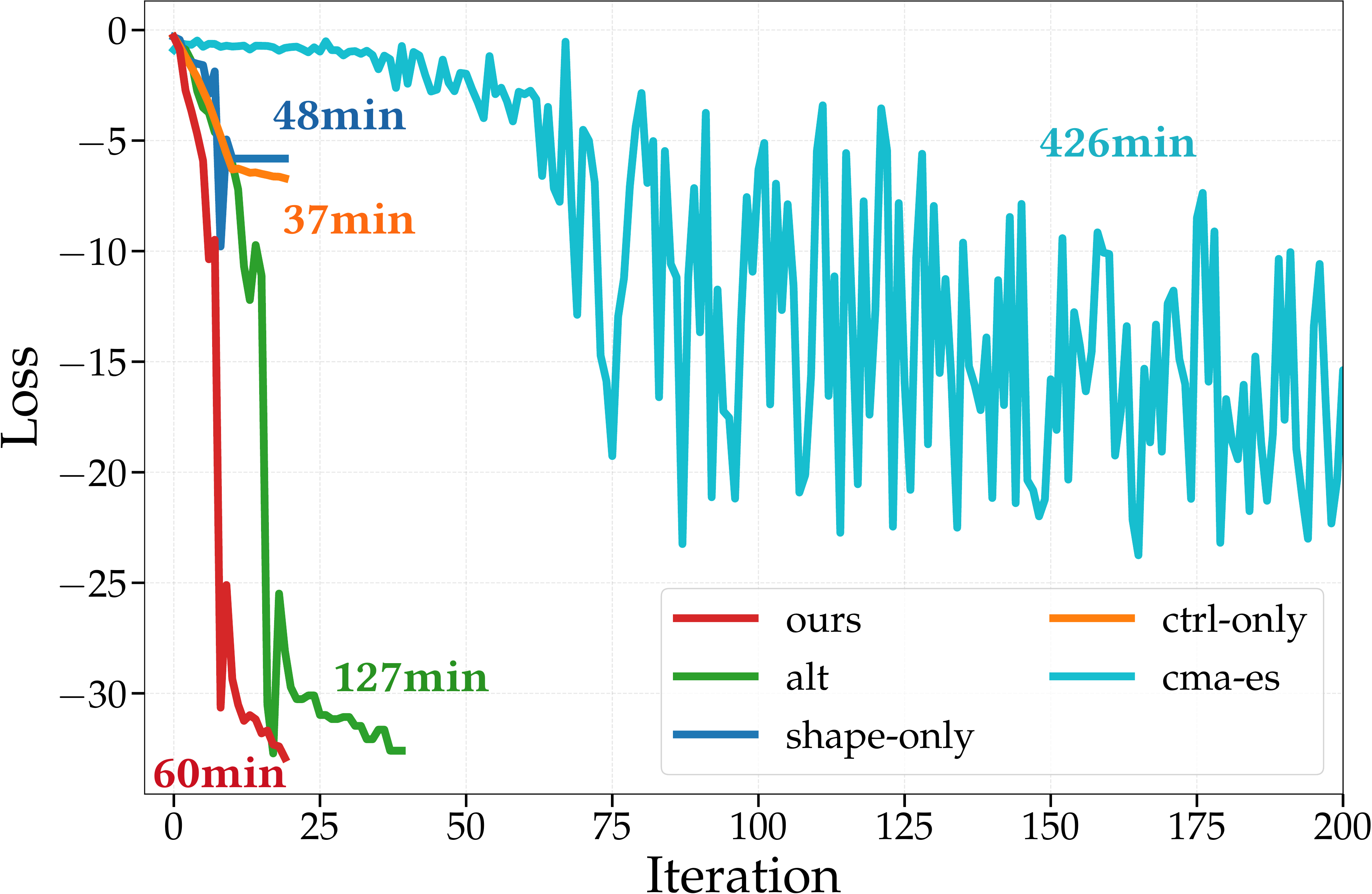}
    \vspace{-1.em}
    \caption{The loss-iteration curves for our method and four baseline algorithms evaluated in the ``open-loop co-optimization'' experiment. Lower losses are better. The CMA-ES loss continually oscillates and does not reach a loss lower than our method before exhausting its computational budget (250 iterations, not shown in the figure). \revise{We label the total time cost for each method with its corresponding color.}}
    \label{fig:exp3_loss}
\end{figure}

For comparison, we also examine the performances of a few baseline algorithms: \textbf{alt}: alternating between shape and control optimization; \textbf{shape-only}: fixing the initial controller and optimizing the shape; \textbf{control-only}: fixing the initial shape and optimizing the controller; \textbf{cma-es}: co-optimizing both shape and controller with CMA-ES \cite{hansen2003reducing}, a gradient-free evolutionary algorithm. By comparing the loss-iteration curves from all of these methods (Fig.~\ref{fig:exp3_loss}), we reach the following conclusions: First, co-optimizing both the shape and controller (\textbf{ours}, \textbf{alt}, and \textbf{cma-es}) reaches a much lower loss than only optimizing shape or control (\textbf{shape-only} and \textbf{control-only}), which is as expected. Second, gradient-based co-optimization (\textbf{ours} and \textbf{alt}) converges significantly faster than the gradient-free baseline \textbf{cma-es}, demonstrating the value of gradients in design optimization. Lastly, we find that simultaneously optimizing both shape and controller (\textbf{ours}) converges to similar results but faster than the alternating strategy $\textbf{alt}$. This observation highlights the value of our differentiable geometric design space, which makes simultaneously co-optimizing shape and control with a gradient-based algorithm possible.

\paragraph{Closed-loop co-optimization} In this experiment, we replace the open-loop controller from the last experiment with a closed-loop neural network controller. We also use the three base shapes in the \reviserm{``shape optimization''} \revise{``shape exploration''} experiment (Fig.~\ref{fig:exp1}) for shape interpolation. The objective is the same as in the previous experiment\reviserm{, \emph{i.e.}, find an optimal shape and a controller that leads to the longest forward travel distance within a given time period}.

\begin{figure}[t]
    \centering
    \includegraphics[width=\linewidth]{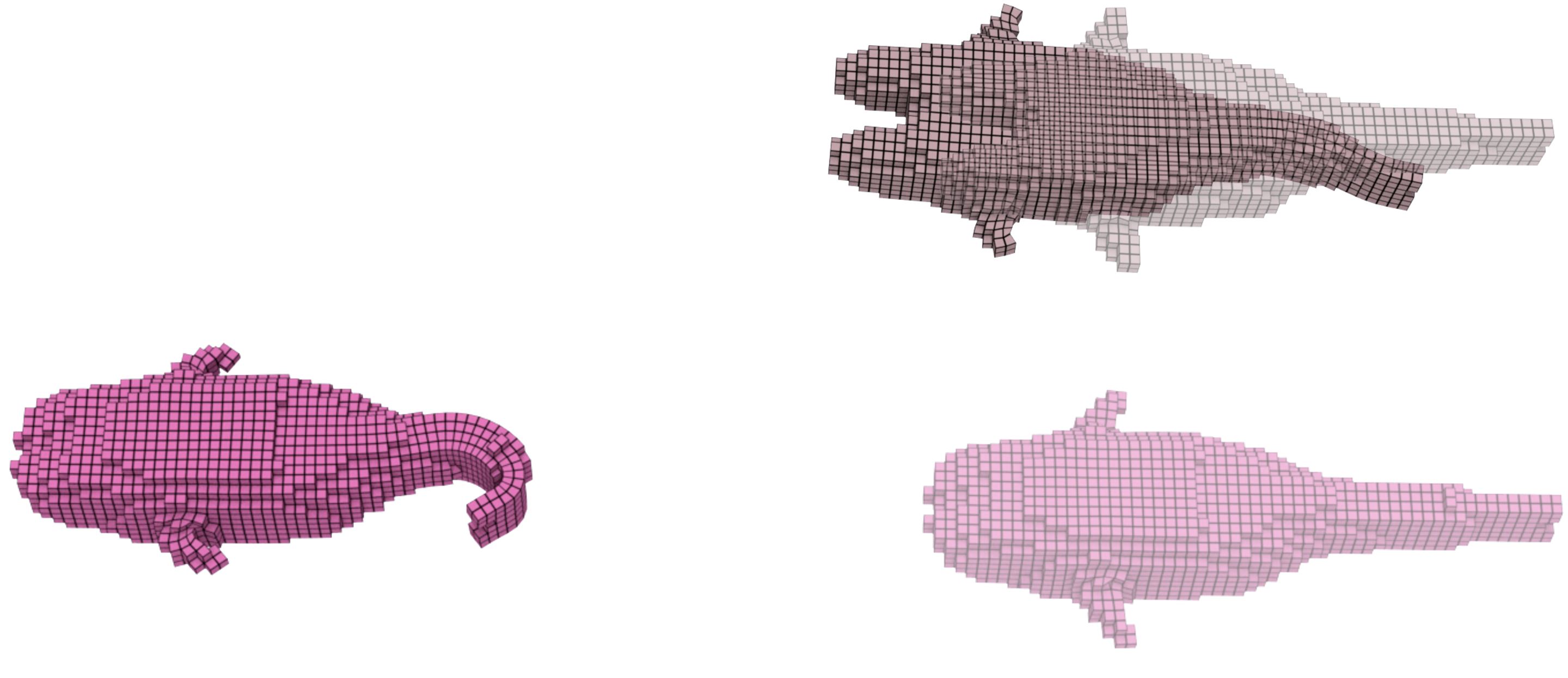}
    \vspace{-2em}
    \caption{``Closed-loop co-optimization'' experiment: Performances of the initial (top) and the optimized geometric design (bottom) of the swimmer. The color of the swimmer is interpolated with their weights $\bm{\alpha}$ from the base shapes' colors in Fig.~\ref{fig:exp1}. The design is simulated to swim forward from an initial position (transparent) on the right. The swimmer's final locations after a fixed amount of time are rendered as solid meshes. A longer traveling distance is preferred in this experiment.}
    \label{fig:exp4_shape}
\end{figure}

\begin{figure*}[htp]
    \centering
    \begin{flushleft}\hspace{14em} {\small Without temporal encoding} \hspace{14em} {\small With temporal encoding}\end{flushleft}\vspace{0em}
    \begin{flushleft}\hspace{12em} {\small Fin  \hspace{12em} Tail  \hspace{13em} Fin  \hspace{12em} Tail}\end{flushleft}
    \hfill
    \includegraphics[width=0.22\linewidth]{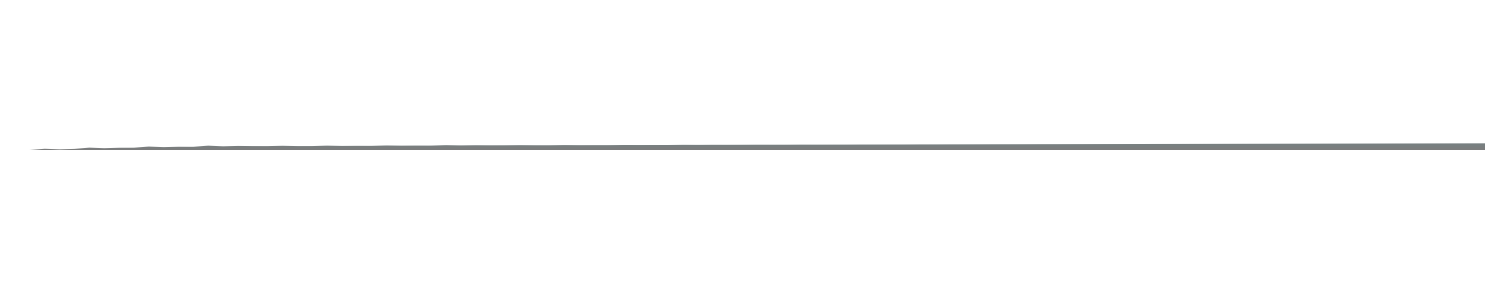}
    \includegraphics[width=0.22\linewidth]{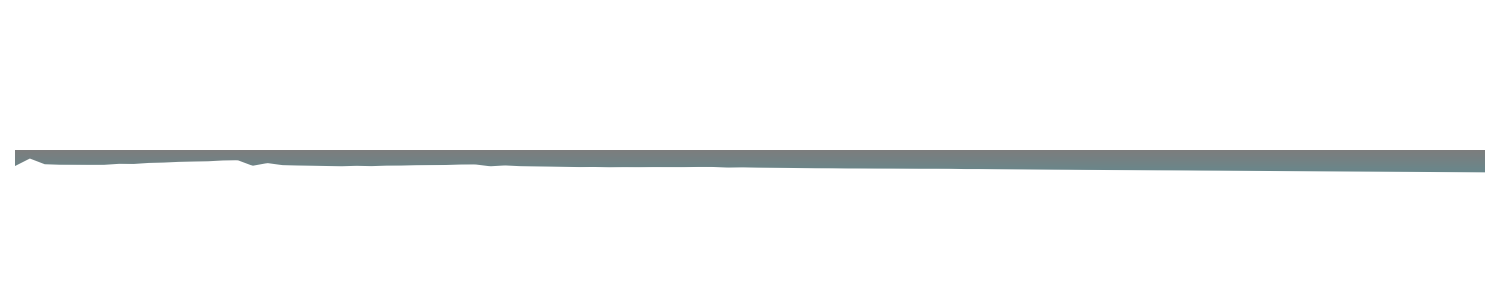}
    \includegraphics[width=0.22\linewidth]{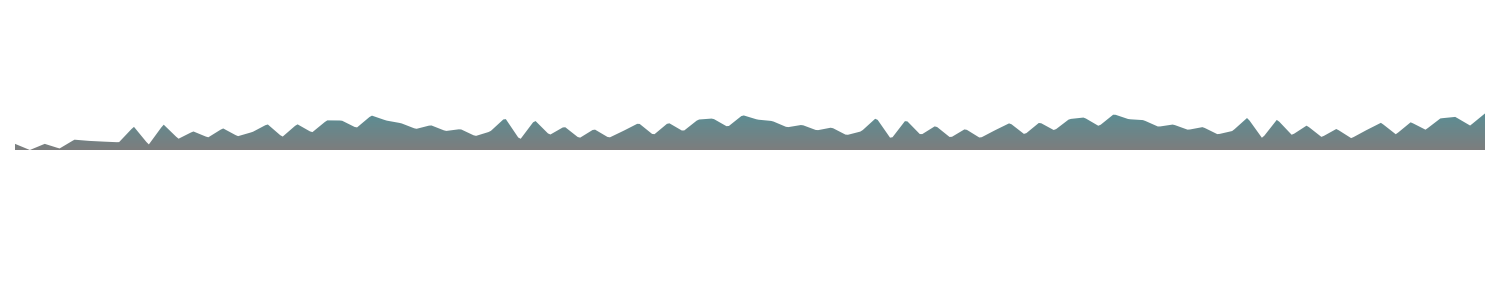}
    \includegraphics[width=0.22\linewidth]{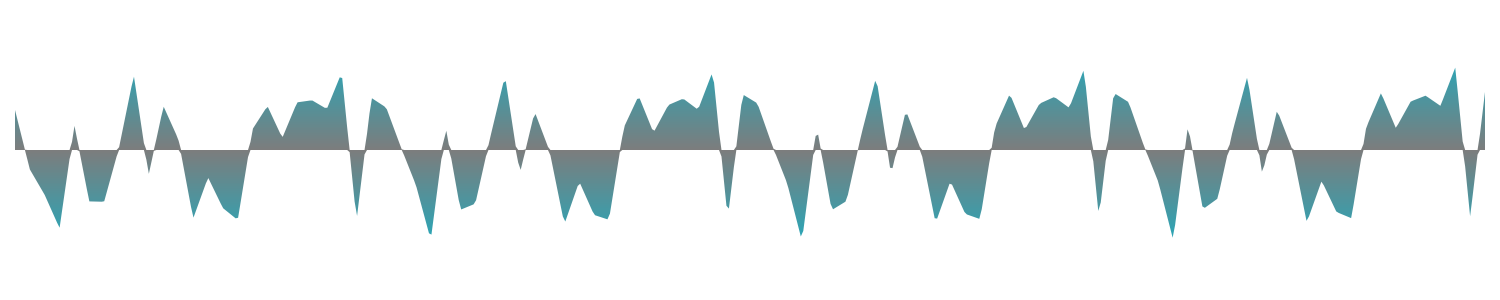}\vspace{0.5em}\\
    \hfill
    \includegraphics[width=0.22\linewidth]{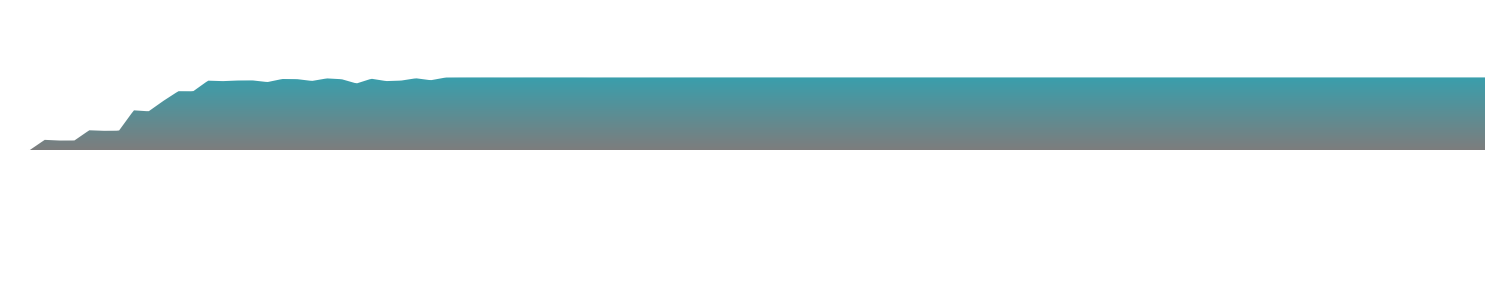}
    \includegraphics[width=0.22\linewidth]{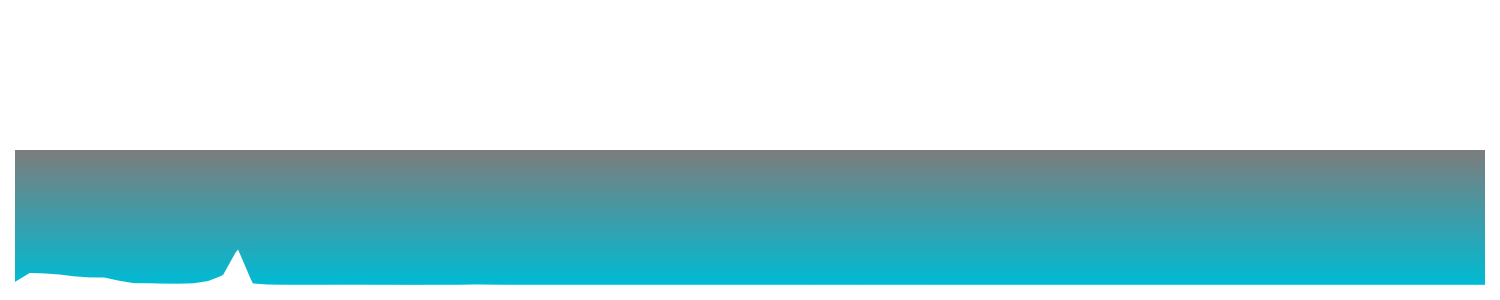}
    \includegraphics[width=0.22\linewidth]{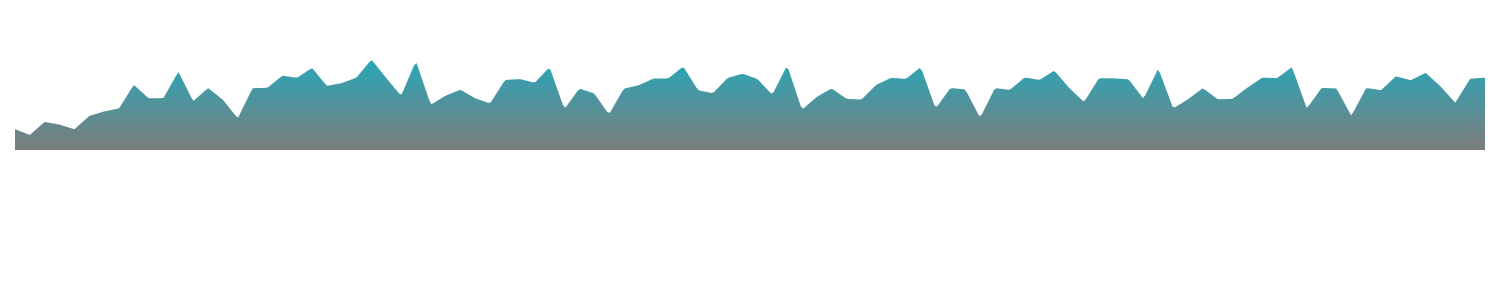}
    \includegraphics[width=0.22\linewidth]{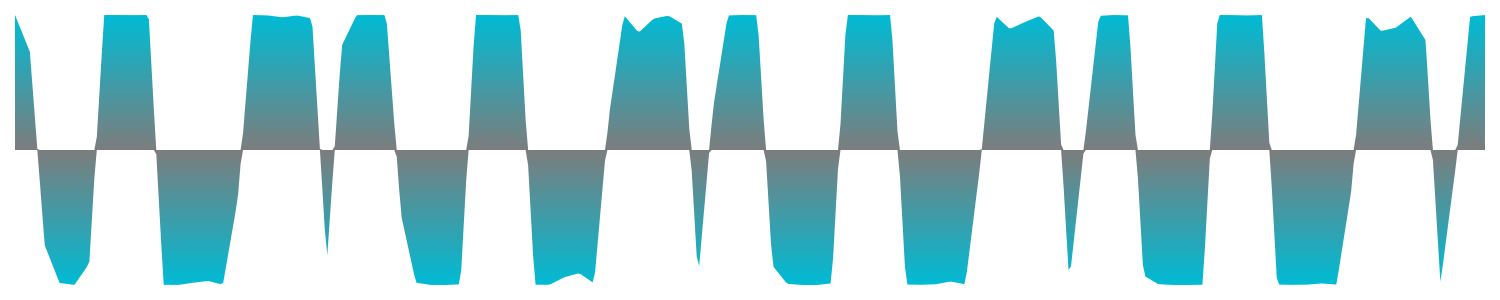}
    \vspace{-6em}
    \begin{flushleft}{\small Initial guess}\end{flushleft}
    \vspace{2em}
    \begin{flushleft}{\small Optimized}\end{flushleft}\vspace{0.5em}
    \caption{The initial and optimized control signals generated by running our method with and without temporal encoding in the ``closed-loop co-optimization'' experiment.}
    \label{fig:exp4_ctrl}
\end{figure*}

\begin{figure}[htp]
    \centering
    \includegraphics[width=\linewidth]{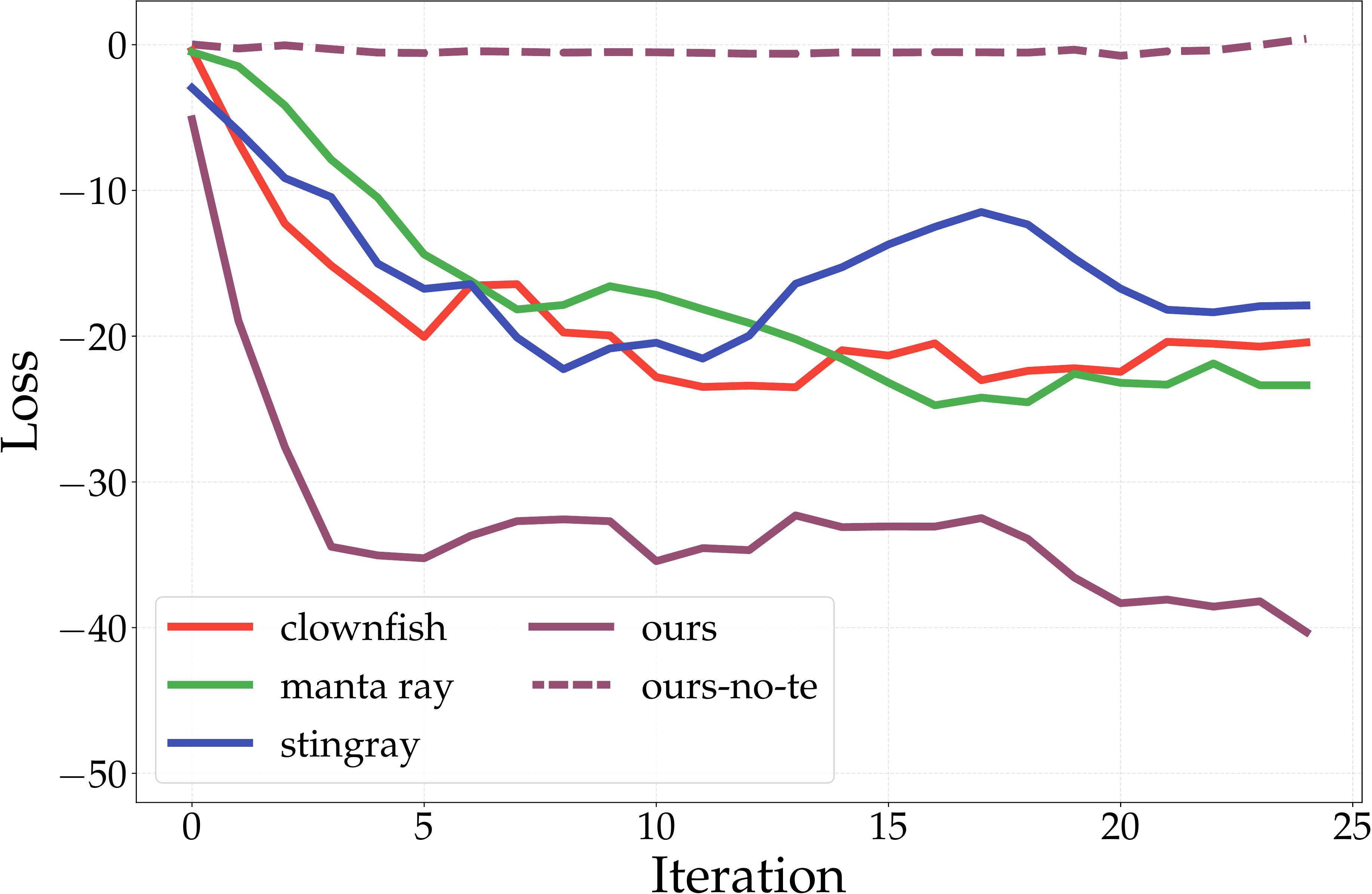}
    \caption{The loss-iteration curves for five different methods in the ``closed-loop co-optimization'' experiment: The curves labeled as ``ours'' and ``ours-no-te'' represent losses from our methods with and without temporal encoding. The ``clownfish'', ``manta ray'', and ``stingray'' curves record the intermediate losses from optimizing controllers with shapes fixed to the corresponding base shape. Lower losses are better.}
    \label{fig:exp4_loss}
\end{figure}

Fig.~\ref{fig:exp4_shape} and Fig.~\ref{fig:exp4_ctrl} show the optimal shape and controller from our method. We observe that the optimal shape assigns more weight to the clownfish and the stingray than the manta ray, likely the clownfish has a larger tail. For the controller, we also notice that the optimal neural network controller discovers a strongly oscillating pattern for the tail (Fig.~\ref{fig:exp4_ctrl}), which aligns with our expectation for a fast swimmer. We point out that the source of periodicity in the control signal is from the temporal encoding technique: As shown in Fig.~\ref{fig:exp4_ctrl}, learning a periodic control output becomes much more difficult when temporal encoding is disabled, leading to much worse performance (Fig.~\ref{fig:exp4_loss}).

To show the value of co-optimizing both shape and control in this experiment (rather than just control), we test fixing the shape design to the three bases and optimizing the controller only. We present the loss-iteration curves for these methods in Fig.~\ref{fig:exp4_loss} and conclude that co-optimizing both the shape and control achieves significantly better performance.

\paragraph{Large dataset co-optimization} In a more realistic example of design pertinent to roboticists, we consider a position-keeping task in the face of an external disturbance from a constant water flow. We feed our shape interpolation a plethora of bases, including four shark variations, seven goldfish variations, and one submarine (Fig.~\ref{fig:teaser}). We again use a closed-loop controller as in the Experiment 4. The swimmer's goal is to maintain its position and orientation in a stream of fast-flowing water moving against its head. Formally, our loss $L$ is defined as follows:
\begin{align}
L=&L_{\text{perf}}+\gamma L_{\text{reg}} \\
L_{\text{perf}}=&\sum_i \|\bm{q}_{i,c}-\bm{q}_{\text{target}}\|_1 \\
L_{\text{reg}}=-&\sum_i (\bm{q}_{i,h_0} - \bm{q}_{i,h_1})\cdot \bm{d}_{\text{target}}.
\end{align}
Here, the loss function consists of a performance loss $L_{\text{perf}}$ and a regularizer $L_{\text{reg}}$. We set the regularizer weight $\gamma=0.01$. The performance loss is defined as the cumulative offset of a node $\bm{q}_{i,c}\in\mathcal{R}^d$ at the $i$-th timestep with respect to a target location $\bm{q}_{\text{target}}=\bm{0}$. Here $c$ is the index of the central node in the aforementioned $Spine$ set. In short, the performance loss encourages the swimmer to stay at the origin within the given time period. Additionally, we introduce the regularizer loss to penalize controllers circling around $\bm{q}_{\text{target}}$. Here, $h_0$ and $h_1$ are indices of two prespecified nodes from the $Spine$ set. Therefore, $\bm{q}_{i,h_0}-\bm{q}_{i,h_1}$ estimates the swimmer's heading at the $i$-th timestep. The regularizer computes the dot product between the true heading and a target heading $\bm{d}_{\text{target}}$, which is the positive $x$ unit vector, to encourage a controller that maintains orientation along the positive $x$ direction.

\begin{figure}[htp]
    \centering
    \includegraphics[width=\linewidth]{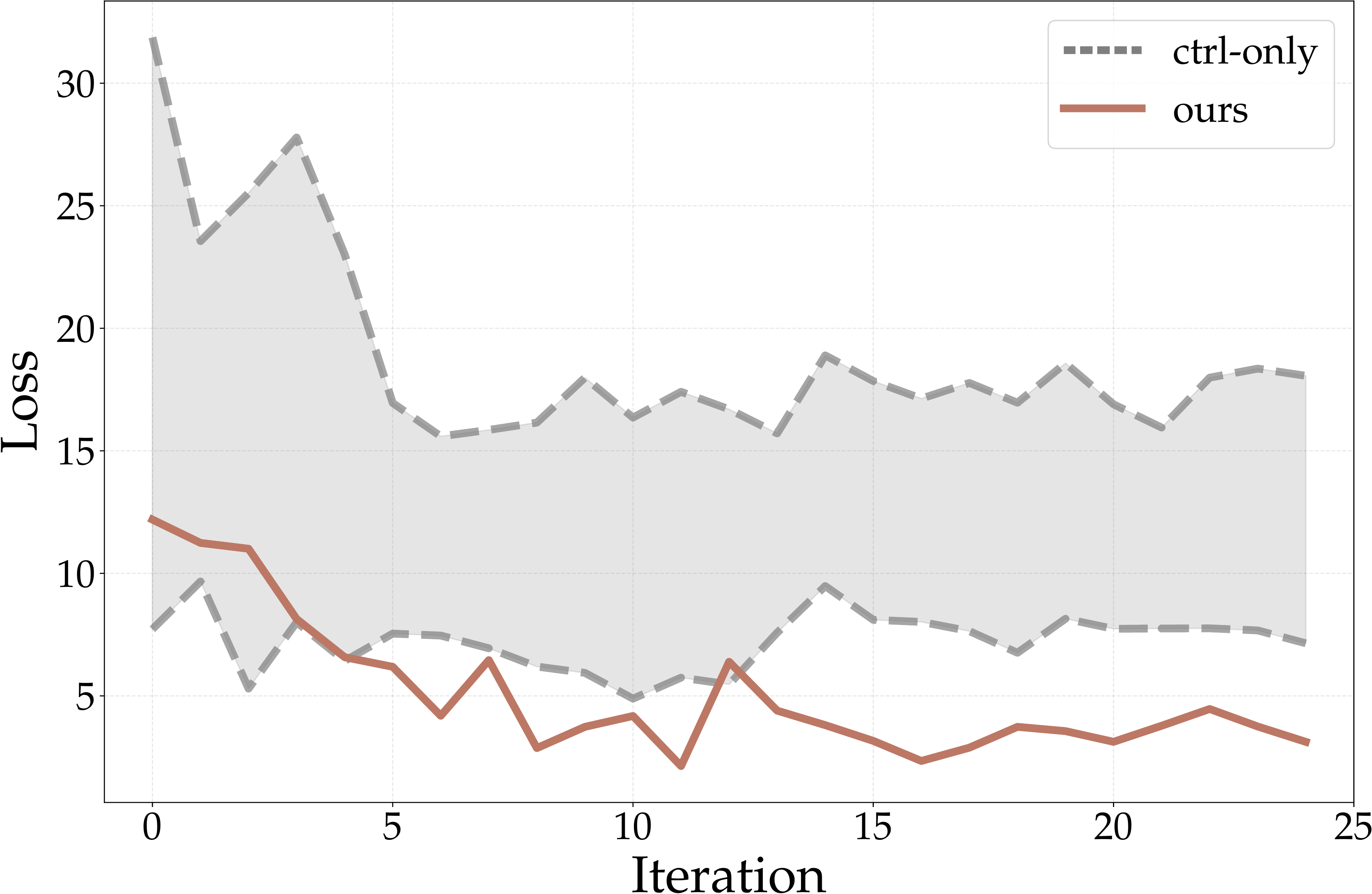}
    \caption{The loss-iteration curves for our method (ours) and optimizing control parameters with the shape fixed to each of the 12 base shapes (ctrl-only). A lower loss is better. We report the aggregate results about the 12 control-only optimizations (whose loss curves all fall in the gray area), with the lower and upper bounds of the loss curves highlighted by dashed lines.}
    \label{fig:exp5}
\end{figure}

We report the optimal shape and controller from our method in Fig.~\ref{fig:teaser}. With  co-optimized shape and control, the swimmer learns to leverage an oscillating motion to counter the flow and stabilize itself. We use an average of all shape bases as the initial shape for optimization. The swimmer's shape after optimization appears mildly different from the initial guess (Fig.~\ref{fig:teaser} middle and right), but that difference has a significant impact on performance.  Further, the difference between the initial and optimized control signals are quite noticeable. In particular, the optimizer learns to intensify the magnitude of the control signal to counter the flow. To justify the importance of shape optimization, we compare our method to optimizing controllers with the shape fixed as each of the 12 bases. As shown in Fig.~\ref{fig:exp5}, our co-optimized swimmer outperforms all 12 base swimmers by a clear margin, highlighting the necessity of co-optimizing both the geometry and the control of the swimmer.

\subsection{Multi-Objective Design}
\paragraph{Multi-objective co-optimization} Finally, we employ our method to investigate a multi-objective design problem: What is the optimal design of a swimmer for both fast \textit{and} efficient forward swimming? These two objectives often conflict with each other for real marine creatures~\cite{sfakiotakis1999review}; therefore, they define a gamut of designs with varying preferences on these two objectives and an interesting Pareto front. 

Formally, we consider the same set of shape bases as in the ``Large dataset co-optimization'' experiment with two loss functions $L_{\text{speed}}$ and $L_{\text{efficiency}}$. We use the loss function in Eqn.\,(\ref{eq:loss_dist}) for $L_{\text{speed}}$, which reaches its minimum when the swimmer obtains the maximum average forward speed over a given time. The efficiency loss is defined as follows:
\begin{align}
L_{\text{efficiency}} = &-\sum_i \frac{|P^{\text{thrust}}|}{1+|P^{\text{thrust}}|+|P^{\text{drag}}|} \\
= &-\sum_i \frac{|\bar{\bm{f}}_i^{\text{thrust}}\cdot\bar{\bm{v}}_i|}{1+|\bar{\bm{f}}_i^{\text{thrust}}\cdot\bar{\bm{v}}_i|+|\bar{\bm{f}}_i^{\text{drag}}\cdot\bar{\bm{v}}_i|}.
\end{align}
In short, $L_{\text{efficiency}}$ provides a measure of the energy dissipation due to the hydrodynamic drag, and minimizing $L_{\text{efficiency}}$ encourages a more efficient usage of the hydrodynamic force. We use $P_{\text{thrust}}$ and $P_{\text{drag}}$ to denote the power of hydrodynamic thrust and drag, respectively. At the $i$-th time step, we define the hydrodynamic force's power $P^{\text{thrust}}$ as the dot product between the average hydrodynamic force $\bar{\bm{f}}_i^{\text{thrust}}$ on the swimmer's surface and the average velocity $\bar{\bm{v}}_i$ computed from nodes in the $Spine$ set. The hydrodynamic drag's power $P^{\text{drag}}$ is defined similarly. Finally, we add $1$ in the denominator to avoid singularities, which occurs when both the water and the swimmer are still.

\begin{figure*}[htp]
    \centering
    \includegraphics[width=\textwidth]{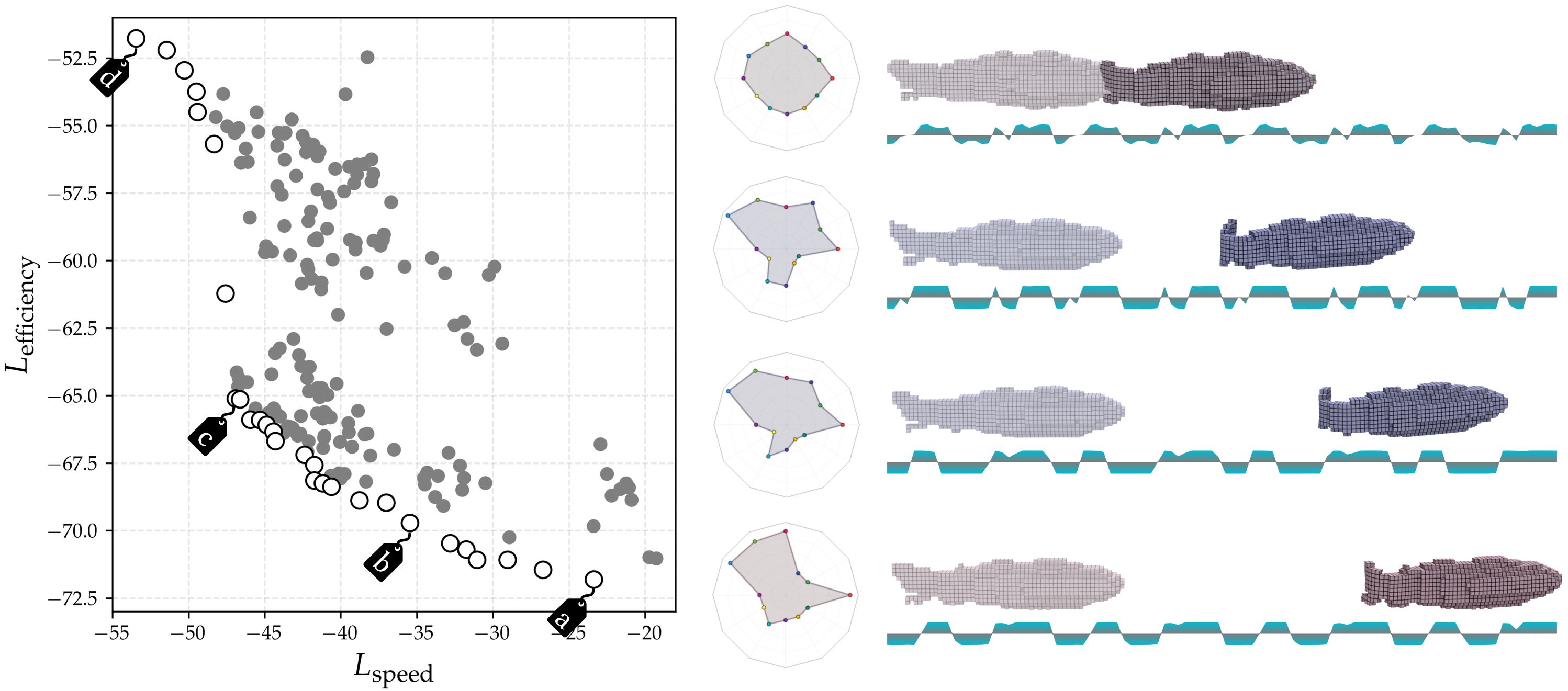}
    \vspace{-25.7em}
    \begin{flushright}{\small (a) high efficiency}\end{flushright}\vspace{3.8em}
    \begin{flushright}{\small (b) medium efficiency}\end{flushright}\vspace{4.2em}
    \begin{flushright}{\small (c) medium speed}\end{flushright}\vspace{4.5em}
    \begin{flushright}{\small (d) high speed}\end{flushright}\vspace{5em}
    \caption{``Multi-objective co-optimization'' experiment: Left: The performance gamut of the intermediate discovered designs (gray dots) and its Pareto front (white circles). Lower losses are better. Right: We show four samples, labeled as (a), (b), (c), and (d), from the Pareto front. For each sample, we show the swimmer's initial (transparent) and final (solid) location at the beginning and end of the simulation. A larger distance between these two locations indicate a faster average speed (better $L_{speed}$). The design parameters for each sample are shown in the radar chart to its left. We plot the control signals for each sample below its initial and final locations.}
    \label{fig:exp6}
\end{figure*}

To understand the implications of different geometry and controller designs on the two losses, we generate a gamut of swimmers and visualize their performances in the $L_{\text{speed}}$-$L_{\text{efficiency}}$ space (Fig.~\ref{fig:exp6} left). Note that lower losses are better, so designs closer to the lower-left corner are preferred. To generate this gamut, we optimize the weighted sum of the two losses $w_s L_{\text{speed}} + w_e L_{\text{efficiency}}$ with $w_s = 0, 0.1, 0.2, \cdots, 1$ and $w_e = 1 - w_s$. We record all intermediate designs discovered during this process to form the gamut, with the designs on the Pareto front highlighted as white circles.

To understand the Pareto optimal designs, we sampled four swimmers along the Pareto front. Fig.~\ref{fig:exp6} further shows the geometric design and neural network control outputs for the sampled swimmers. We find that the four swimmers' shape designs are quite similar, but their controllers are significantly different. In particular, their control signals show a strong correlation with the losses: controllers preferred by faster swimmers show larger magnitudes lasting for a longer period, which exerting more powerful forces from the muscle fibers in the actuators. These sampled four swimmers, along with many others discovered in the Pareto front, form a diverse set of designs for users to choose from in case they have varying preferences.

\subsection{Ablation Study}\label{sec:subsec:ab_study}

\begin{figure}[tp]
    \centering
    \begin{subfigure}[b]{0.49\linewidth}
        \includegraphics[width=\linewidth]{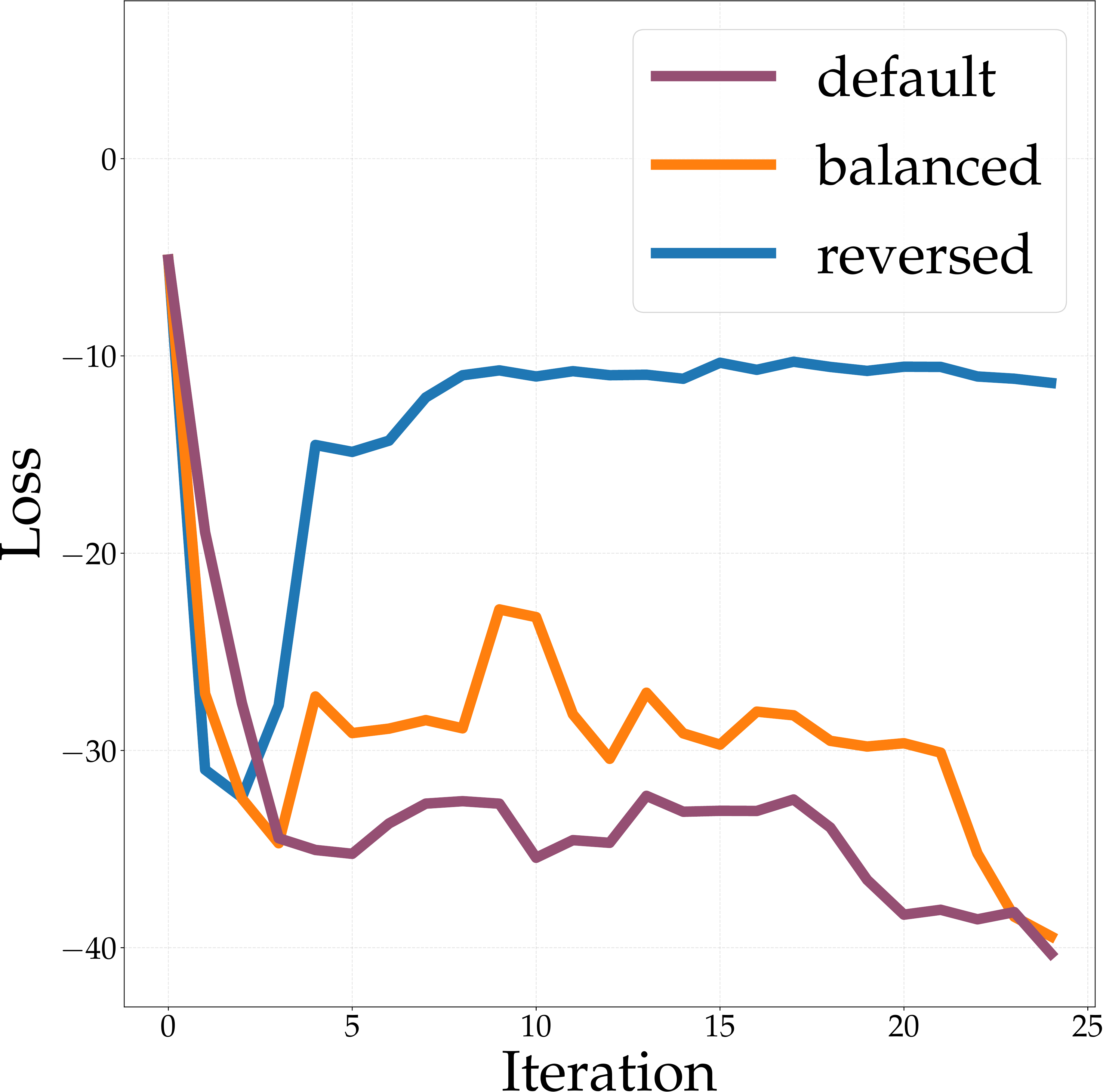}
    \end{subfigure}
    \begin{subfigure}[b]{0.49\linewidth}
        \includegraphics[width=\linewidth]{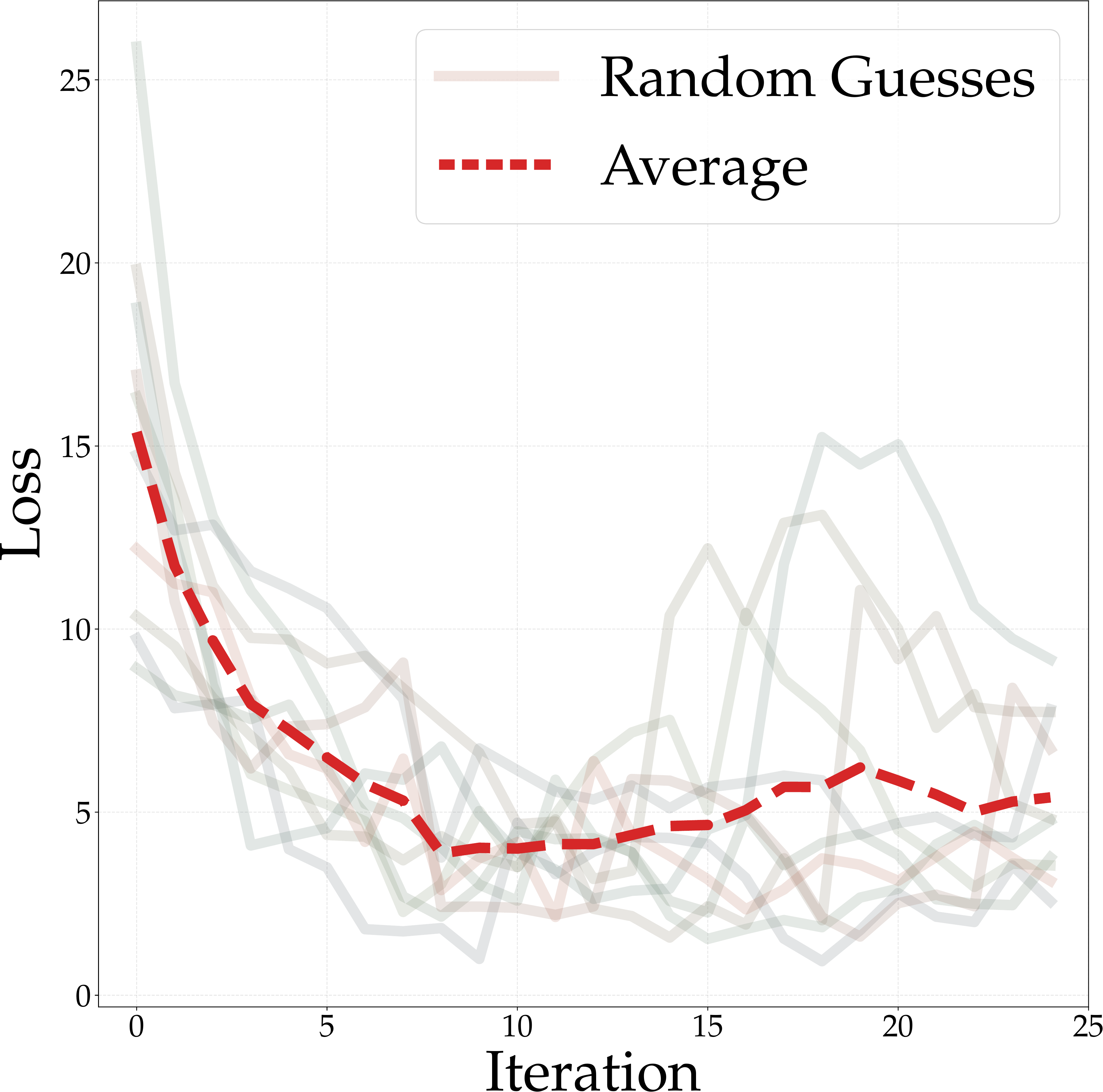}
    \end{subfigure}
    \caption{\revise{The loss-iteration curves for ablation studies on the ``gradient scaling'' experiment (left) and the ``initial guesses'' experiment (right) in Sec.~\ref{sec:subsec:ab_study}. Lower losses are better. Left: The ``default'' curve reports the result without scaling the gradients. The ``balanced'' and ``reversed'' curves represent the strategies normalizing the gradients to the same average $L1$-norm and switching the scales, respectively. Right: The semitransparent curves represent the losses from different initial guesses. The red dashed line reports the average loss.}}
    \label{fig:hyper}
\end{figure}

\revise{\paragraph{Gradient Scaling}
The Adam optimizer we use in our experiments is a first-order gradient descent optimizer. Since such algorithms are generally not scale-invariant and we co-optimize parameters from two very different categories (geometry and control), the possible imbalance between the scale of geometry and control parameters may affect the optimizer's performance. To examine the impact of different scales, we rerun the ``closed-loop co-optimization'' experiment by scaling the geometry and control parameters in three different settings: First, the default setting repeats the experiment with no changes, in which case we notice the ratio between the gradient from each geometry and control parameter is roughly 3:1. Next, in the balanced setting, we rescale the control parameters by roughly a factor of 3 so that their gradients have magnitudes comparable to those from the geometry parameters. Finally, in the reversed setting, we further rescale the control parameters until the ratio between the geometry and control gradients becomes 1:3, the reciprocal of the ratio in the default setting. We report the training curves in Fig.~\ref{fig:hyper} (left), from which we notice a substantial effect from rescaling these parameters as expected. Using a scale-invariant optimizer instead of Adam could be a potential solution in the future.
}

\revise{\paragraph{Initial Guesses} To better understand the influence of different initial guesses on the performance of our method, we rerun the ``large dataset co-optimization'' experiment with ten randomly sampled initial shapes and controllers. Fig.~\ref{fig:hyper} (right) reports the resulting ten training curves, from which we observe a consistent decrease in the loss function across all initial guesses. Still, we notice that not all random guesses converge to the same optimal solution, and some of them are trapped in different local minima. Such results are not surprising as our gradient-based method is inherently a continuous local optimization algorithm. More advanced global search algorithms might alleviate the issue of local minima, which we consider as future work.
}

\section{Conclusions, Limitations, and Future Work}
We have presented a method for co-optimizing soft swimmers over control and complex geometry.  By exploiting differentiable simulation and control and a basis space governed by the Wasserstein distance, we are able to generate biomimetic forms that can swim quickly, resist disturbance, or save energy. Further, we have generated designs that are Pareto-optimal in two conflicting design objectives.  Our co-optimization procedure outperforms optimizing over each domain independently, demonstrating the tight interrelation of form and control in swimmer behavior.

While we have provided a first foray into the computational design of soft underwater swimmers, a number of interesting problems remain ripe for exploration.
First, while our algorithm was able to interpolate between actuator shapes, those actuators were placed manually on the base shapes.  A method for automating the design of muscle-based actuators for soft swimmers would be interesting.
Second, our algorithm requires that the base shapes themselves be chosen by hand --- it would be interesting to investigate methods for extending the morphological search beyond the Wasserstein basis space. \revise{For example, it can be combined with discrete, composition-based design methods, \emph{e.g.}, \citet{zhao2020}, to explore combinatorial design space.}
Third, certain aspects of soft swimmer design --- \emph{e.g.} sensing and material selection --- were untouched in this work.
Fourth, our simulator can be made more physically realistic by handling environmental contact as well as employing computational fluid dynamics rather than our analytical hydrodynamic model.
Fifth, our work presented here investigated only virtual swimmers.  It would be interesting to fabricate their physical counterparts, and research methods for overcoming the likely sim-to-real gap for physical soft swimmers.
\revise{Sixth, the objectives of optimization in our experiments cover only travel distance, the ability for position maintenance, and swimming efficiency. It will be exciting to extend our algorithm to more complex goals, \emph{e.g.}, stability under non-constant current or controllability over a target trajectory.}
Finally, as our optimization scheme is a local, gradient-based method, there is no guarantee for global optimality.  It would be interesting to see if our gradient-based optimization could be combined with more global heuristic searches (such as simulated annealing or evolutionary algorithms) to reap the benefits of both approaches.

\begin{acks}
We thank Yue Wang for the valuable discussion on the Wasserstein barycentric interpolation. We also thank the anonymous reviewers for their constructive comments. This work is supported by Intelligence Advanced Research Projects Agency (grant No.~2019-19020100001) and Defense Advanced Research Projects Agency (grant No.~FA8750-20-C-0075).
\end{acks}

\bibliographystyle{ACM-Reference-Format}
\bibliography{main}


\begin{thebibliography}{74}


\ifx \showCODEN    \undefined \def \showCODEN     #1{\unskip}     \fi
\ifx \showDOI      \undefined \def \showDOI       #1{#1}\fi
\ifx \showISBNx    \undefined \def \showISBNx     #1{\unskip}     \fi
\ifx \showISBNxiii \undefined \def \showISBNxiii  #1{\unskip}     \fi
\ifx \showISSN     \undefined \def \showISSN      #1{\unskip}     \fi
\ifx \showLCCN     \undefined \def \showLCCN      #1{\unskip}     \fi
\ifx \shownote     \undefined \def \shownote      #1{#1}          \fi
\ifx \showarticletitle \undefined \def \showarticletitle #1{#1}   \fi
\ifx \showURL      \undefined \def \showURL       {\relax}        \fi
\providecommand\bibfield[2]{#2}
\providecommand\bibinfo[2]{#2}
\providecommand\natexlab[1]{#1}
\providecommand\showeprint[2][]{arXiv:#2}

\bibitem[\protect\citeauthoryear{Averkiou, Kim, Zheng, and Mitra}{Averkiou
  et~al\mbox{.}}{2014}]%
        {averkiou2014shapesynth}
\bibfield{author}{\bibinfo{person}{Melinos Averkiou},
  \bibinfo{person}{Vladimir~G Kim}, \bibinfo{person}{Youyi Zheng}, {and}
  \bibinfo{person}{Niloy~J Mitra}.} \bibinfo{year}{2014}\natexlab{}.
\newblock \showarticletitle{Shapesynth: Parameterizing model collections for
  coupled shape exploration and synthesis}. In
  \bibinfo{booktitle}{\emph{Computer Graphics Forum}},
  Vol.~\bibinfo{volume}{33}. Wiley Online Library, \bibinfo{pages}{125--134}.
\newblock


\bibitem[\protect\citeauthoryear{Baek, Lim, and Lee}{Baek
  et~al\mbox{.}}{2015}]%
        {baek2015isometric}
\bibfield{author}{\bibinfo{person}{Seung-Yeob Baek}, \bibinfo{person}{Jeonghun
  Lim}, {and} \bibinfo{person}{Kunwoo Lee}.} \bibinfo{year}{2015}\natexlab{}.
\newblock \showarticletitle{Isometric shape interpolation}.
\newblock \bibinfo{journal}{\emph{Computers \& Graphics}}  \bibinfo{volume}{46}
  (\bibinfo{year}{2015}), \bibinfo{pages}{257--263}.
\newblock


\bibitem[\protect\citeauthoryear{Barbi{\v{c}}, da~Silva, and
  Popovi{\'c}}{Barbi{\v{c}} et~al\mbox{.}}{2009}]%
        {barbivc2009deformable}
\bibfield{author}{\bibinfo{person}{Jernej Barbi{\v{c}}}, \bibinfo{person}{Marco
  da Silva}, {and} \bibinfo{person}{Jovan Popovi{\'c}}.}
  \bibinfo{year}{2009}\natexlab{}.
\newblock \showarticletitle{Deformable object animation using reduced optimal
  control}.
\newblock \bibinfo{journal}{\emph{ACM Transactions on Graphics (TOG)}}
  \bibinfo{volume}{28}, \bibinfo{number}{3} (\bibinfo{year}{2009}),
  \bibinfo{pages}{1--9}.
\newblock


\bibitem[\protect\citeauthoryear{Barbi{\v{c}} and Popovi{\'c}}{Barbi{\v{c}} and
  Popovi{\'c}}{2008}]%
        {barbivc2008real}
\bibfield{author}{\bibinfo{person}{Jernej Barbi{\v{c}}} {and}
  \bibinfo{person}{Jovan Popovi{\'c}}.} \bibinfo{year}{2008}\natexlab{}.
\newblock \showarticletitle{Real-time control of physically based simulations
  using gentle forces}.
\newblock \bibinfo{journal}{\emph{ACM transactions on graphics (TOG)}}
  \bibinfo{volume}{27}, \bibinfo{number}{5} (\bibinfo{year}{2008}),
  \bibinfo{pages}{1--10}.
\newblock


\bibitem[\protect\citeauthoryear{Belbute-Peres, Smith, Allen, Tenenbaum, and
  Kolter}{Belbute-Peres et~al\mbox{.}}{2018}]%
        {de2018end}
\bibfield{author}{\bibinfo{person}{Filipe de~A Belbute-Peres},
  \bibinfo{person}{Kevin~A Smith}, \bibinfo{person}{Kelsey~R Allen},
  \bibinfo{person}{Joshua~B Tenenbaum}, {and} \bibinfo{person}{J~Zico Kolter}.}
  \bibinfo{year}{2018}\natexlab{}.
\newblock \showarticletitle{End-to-end differentiable physics for learning and
  control}. In \bibinfo{booktitle}{\emph{Proceedings of the 32nd International
  Conference on Neural Information Processing Systems}}.
  \bibinfo{pages}{7178--7189}.
\newblock


\bibitem[\protect\citeauthoryear{Berlinger, Duduta, Gloria, Clarke, Nagpal, and
  Wood}{Berlinger et~al\mbox{.}}{2018}]%
        {berlinger2018modular}
\bibfield{author}{\bibinfo{person}{Florian Berlinger}, \bibinfo{person}{Mihai
  Duduta}, \bibinfo{person}{Hudson Gloria}, \bibinfo{person}{David Clarke},
  \bibinfo{person}{Radhika Nagpal}, {and} \bibinfo{person}{Robert Wood}.}
  \bibinfo{year}{2018}\natexlab{}.
\newblock \showarticletitle{A modular dielectric elastomer actuator to drive
  miniature autonomous underwater vehicles}. In \bibinfo{booktitle}{\emph{2018
  IEEE International Conference on Robotics and Automation (ICRA)}}. IEEE,
  \bibinfo{pages}{3429--3435}.
\newblock


\bibitem[\protect\citeauthoryear{Bonneel, Peyr{\'e}, and Cuturi}{Bonneel
  et~al\mbox{.}}{2016}]%
        {bonneel2016wasserstein}
\bibfield{author}{\bibinfo{person}{Nicolas Bonneel}, \bibinfo{person}{Gabriel
  Peyr{\'e}}, {and} \bibinfo{person}{Marco Cuturi}.}
  \bibinfo{year}{2016}\natexlab{}.
\newblock \showarticletitle{Wasserstein barycentric coordinates: histogram
  regression using optimal transport}.
\newblock \bibinfo{journal}{\emph{ACM Transactions on Graphics (TOG)}}
  \bibinfo{volume}{35}, \bibinfo{number}{4} (\bibinfo{year}{2016}),
  \bibinfo{pages}{1--10}.
\newblock


\bibitem[\protect\citeauthoryear{Bronstein, Bronstein, Guibas, and
  Ovsjanikov}{Bronstein et~al\mbox{.}}{2011}]%
        {bronstein2011shape}
\bibfield{author}{\bibinfo{person}{Alexander~M Bronstein},
  \bibinfo{person}{Michael~M Bronstein}, \bibinfo{person}{Leonidas~J Guibas},
  {and} \bibinfo{person}{Maks Ovsjanikov}.} \bibinfo{year}{2011}\natexlab{}.
\newblock \showarticletitle{Shape google: Geometric words and expressions for
  invariant shape retrieval}.
\newblock \bibinfo{journal}{\emph{ACM Transactions on Graphics (TOG)}}
  \bibinfo{volume}{30}, \bibinfo{number}{1} (\bibinfo{year}{2011}),
  \bibinfo{pages}{1--20}.
\newblock


\bibitem[\protect\citeauthoryear{Chen, Rubanova, Bettencourt, and
  Duvenaud}{Chen et~al\mbox{.}}{2018}]%
        {chen2018neural}
\bibfield{author}{\bibinfo{person}{Ricky~TQ Chen}, \bibinfo{person}{Yulia
  Rubanova}, \bibinfo{person}{Jesse Bettencourt}, {and} \bibinfo{person}{David
  Duvenaud}.} \bibinfo{year}{2018}\natexlab{}.
\newblock \showarticletitle{Neural ordinary differential equations}. In
  \bibinfo{booktitle}{\emph{Proceedings of the 32nd International Conference on
  Neural Information Processing Systems}}. \bibinfo{pages}{6572--6583}.
\newblock


\bibitem[\protect\citeauthoryear{Cheney, MacCurdy, Clune, and Lipson}{Cheney
  et~al\mbox{.}}{2014}]%
        {cheney2014unshackling}
\bibfield{author}{\bibinfo{person}{Nick Cheney}, \bibinfo{person}{Robert
  MacCurdy}, \bibinfo{person}{Jeff Clune}, {and} \bibinfo{person}{Hod Lipson}.}
  \bibinfo{year}{2014}\natexlab{}.
\newblock \showarticletitle{Unshackling evolution: evolving soft robots with
  multiple materials and a powerful generative encoding}.
\newblock \bibinfo{journal}{\emph{ACM SIGEVOlution}} \bibinfo{volume}{7},
  \bibinfo{number}{1} (\bibinfo{year}{2014}), \bibinfo{pages}{11--23}.
\newblock


\bibitem[\protect\citeauthoryear{Corucci, Cheney, Lipson, Laschi, and
  Bongard}{Corucci et~al\mbox{.}}{2016}]%
        {corucci2016evolving}
\bibfield{author}{\bibinfo{person}{Francesco Corucci}, \bibinfo{person}{Nick
  Cheney}, \bibinfo{person}{Hod Lipson}, \bibinfo{person}{Cecilia Laschi},
  {and} \bibinfo{person}{Josh Bongard}.} \bibinfo{year}{2016}\natexlab{}.
\newblock \showarticletitle{Evolving swimming soft-bodied creatures}. In
  \bibinfo{booktitle}{\emph{ALIFE XV, The Fifteenth International Conference on
  the Synthesis and Simulation of Living Systems, Late Breaking Proceedings}},
  Vol.~\bibinfo{volume}{6}.
\newblock


\bibitem[\protect\citeauthoryear{Degrave, Hermans, Dambre, and wyffels}{Degrave
  et~al\mbox{.}}{2019}]%
        {degrave2019differentiable}
\bibfield{author}{\bibinfo{person}{Jonas Degrave}, \bibinfo{person}{Michiel
  Hermans}, \bibinfo{person}{Joni Dambre}, {and} \bibinfo{person}{Francis
  wyffels}.} \bibinfo{year}{2019}\natexlab{}.
\newblock \showarticletitle{A Differentiable Physics Engine for Deep Learning
  in Robotics}.
\newblock \bibinfo{journal}{\emph{Frontiers in Neurorobotics}}
  \bibinfo{volume}{13} (\bibinfo{year}{2019}), \bibinfo{pages}{6}.
\newblock
\showISSN{1662-5218}


\bibitem[\protect\citeauthoryear{Della~Santina, Katzschmann, Biechi, and
  Rus}{Della~Santina et~al\mbox{.}}{2018}]%
        {della2018dynamic}
\bibfield{author}{\bibinfo{person}{Cosimo Della~Santina},
  \bibinfo{person}{Robert~K Katzschmann}, \bibinfo{person}{Antonio Biechi},
  {and} \bibinfo{person}{Daniela Rus}.} \bibinfo{year}{2018}\natexlab{}.
\newblock \showarticletitle{Dynamic control of soft robots interacting with the
  environment}. In \bibinfo{booktitle}{\emph{2018 IEEE International Conference
  on Soft Robotics (RoboSoft)}}. IEEE, \bibinfo{pages}{46--53}.
\newblock


\bibitem[\protect\citeauthoryear{Du, Schulz, Zhu, Bickel, and Matusik}{Du
  et~al\mbox{.}}{2016}]%
        {du2016}
\bibfield{author}{\bibinfo{person}{Tao Du}, \bibinfo{person}{Adriana Schulz},
  \bibinfo{person}{Bo Zhu}, \bibinfo{person}{Bernd Bickel}, {and}
  \bibinfo{person}{Wojciech Matusik}.} \bibinfo{year}{2016}\natexlab{}.
\newblock \showarticletitle{Computational multicopter design}.
\newblock \bibinfo{journal}{\emph{ACM Transactions on Graphics (TOG)}}
  \bibinfo{volume}{35}, \bibinfo{number}{6} (\bibinfo{year}{2016}),
  \bibinfo{pages}{1--10}.
\newblock


\bibitem[\protect\citeauthoryear{Du, Wu, Ma, Wah, Spielberg, Rus, and
  Matusik}{Du et~al\mbox{.}}{2021}]%
        {du2021diffpd}
\bibfield{author}{\bibinfo{person}{Tao Du}, \bibinfo{person}{Kui Wu},
  \bibinfo{person}{Pingchuan Ma}, \bibinfo{person}{Sebastien Wah},
  \bibinfo{person}{Andrew Spielberg}, \bibinfo{person}{Daniela Rus}, {and}
  \bibinfo{person}{Wojciech Matusik}.} \bibinfo{year}{2021}\natexlab{}.
\newblock \showarticletitle{DiffPD: Differentiable Projective Dynamics with
  Contact}.
\newblock \bibinfo{journal}{\emph{arXiv preprint arXiv:2101.05917}}
  (\bibinfo{year}{2021}).
\newblock


\bibitem[\protect\citeauthoryear{Fish and Lauder}{Fish and Lauder}{2006}]%
        {fish2006passive}
\bibfield{author}{\bibinfo{person}{FE Fish} {and} \bibinfo{person}{George~V
  Lauder}.} \bibinfo{year}{2006}\natexlab{}.
\newblock \showarticletitle{Passive and active flow control by swimming fishes
  and mammals}.
\newblock \bibinfo{journal}{\emph{Annu. Rev. Fluid Mech.}}
  \bibinfo{volume}{38} (\bibinfo{year}{2006}), \bibinfo{pages}{193--224}.
\newblock


\bibitem[\protect\citeauthoryear{Fish, Averkiou, Van~Kaick, Sorkine-Hornung,
  Cohen-Or, and Mitra}{Fish et~al\mbox{.}}{2014}]%
        {fish2014meta}
\bibfield{author}{\bibinfo{person}{Noa Fish}, \bibinfo{person}{Melinos
  Averkiou}, \bibinfo{person}{Oliver Van~Kaick}, \bibinfo{person}{Olga
  Sorkine-Hornung}, \bibinfo{person}{Daniel Cohen-Or}, {and}
  \bibinfo{person}{Niloy~J Mitra}.} \bibinfo{year}{2014}\natexlab{}.
\newblock \showarticletitle{Meta-representation of shape families}.
\newblock \bibinfo{journal}{\emph{ACM Transactions on Graphics (TOG)}}
  \bibinfo{volume}{33}, \bibinfo{number}{4} (\bibinfo{year}{2014}),
  \bibinfo{pages}{1--11}.
\newblock


\bibitem[\protect\citeauthoryear{Geijtenbeek, Van De~Panne, and Van
  Der~Stappen}{Geijtenbeek et~al\mbox{.}}{2013}]%
        {geijtenbeek2013flexible}
\bibfield{author}{\bibinfo{person}{Thomas Geijtenbeek},
  \bibinfo{person}{Michiel Van De~Panne}, {and} \bibinfo{person}{A~Frank Van
  Der~Stappen}.} \bibinfo{year}{2013}\natexlab{}.
\newblock \showarticletitle{Flexible muscle-based locomotion for bipedal
  creatures}.
\newblock \bibinfo{journal}{\emph{ACM Transactions on Graphics (TOG)}}
  \bibinfo{volume}{32}, \bibinfo{number}{6} (\bibinfo{year}{2013}),
  \bibinfo{pages}{1--11}.
\newblock


\bibitem[\protect\citeauthoryear{Geilinger, Hahn, Zehnder, B{\"a}cher,
  Thomaszewski, and Coros}{Geilinger et~al\mbox{.}}{2020}]%
        {geilinger2020add}
\bibfield{author}{\bibinfo{person}{Moritz Geilinger}, \bibinfo{person}{David
  Hahn}, \bibinfo{person}{Jonas Zehnder}, \bibinfo{person}{Moritz B{\"a}cher},
  \bibinfo{person}{Bernhard Thomaszewski}, {and} \bibinfo{person}{Stelian
  Coros}.} \bibinfo{year}{2020}\natexlab{}.
\newblock \showarticletitle{ADD: analytically differentiable dynamics for
  multi-body systems with frictional contact}.
\newblock \bibinfo{journal}{\emph{ACM Transactions on Graphics (TOG)}}
  \bibinfo{volume}{39}, \bibinfo{number}{6} (\bibinfo{year}{2020}),
  \bibinfo{pages}{1--15}.
\newblock


\bibitem[\protect\citeauthoryear{Giftthaler, Neunert, St{\"a}uble, Frigerio,
  Semini, and Buchli}{Giftthaler et~al\mbox{.}}{2017}]%
        {giftthaler2017automatic}
\bibfield{author}{\bibinfo{person}{Markus Giftthaler}, \bibinfo{person}{Michael
  Neunert}, \bibinfo{person}{Markus St{\"a}uble}, \bibinfo{person}{Marco
  Frigerio}, \bibinfo{person}{Claudio Semini}, {and} \bibinfo{person}{Jonas
  Buchli}.} \bibinfo{year}{2017}\natexlab{}.
\newblock \showarticletitle{Automatic differentiation of rigid body dynamics
  for optimal control and estimation}.
\newblock \bibinfo{journal}{\emph{Advanced Robotics}} \bibinfo{volume}{31},
  \bibinfo{number}{22} (\bibinfo{year}{2017}), \bibinfo{pages}{1225--1237}.
\newblock


\bibitem[\protect\citeauthoryear{Grzeszczuk and Terzopoulos}{Grzeszczuk and
  Terzopoulos}{1995}]%
        {grzeszczuk1995automated}
\bibfield{author}{\bibinfo{person}{Radek Grzeszczuk} {and}
  \bibinfo{person}{Demetri Terzopoulos}.} \bibinfo{year}{1995}\natexlab{}.
\newblock \showarticletitle{Automated learning of muscle-actuated locomotion
  through control abstraction}. In \bibinfo{booktitle}{\emph{Proceedings of the
  22nd annual conference on Computer graphics and interactive techniques}}.
  \bibinfo{pages}{63--70}.
\newblock


\bibitem[\protect\citeauthoryear{Ha}{Ha}{2019}]%
        {ha2019reinforcement}
\bibfield{author}{\bibinfo{person}{David Ha}.} \bibinfo{year}{2019}\natexlab{}.
\newblock \showarticletitle{Reinforcement learning for improving agent design}.
\newblock \bibinfo{journal}{\emph{Artificial life}} \bibinfo{volume}{25},
  \bibinfo{number}{4} (\bibinfo{year}{2019}), \bibinfo{pages}{352--365}.
\newblock


\bibitem[\protect\citeauthoryear{Ha, Coros, Alspach, Bern, Kim, and Yamane}{Ha
  et~al\mbox{.}}{2018}]%
        {ha2018computational}
\bibfield{author}{\bibinfo{person}{Sehoon Ha}, \bibinfo{person}{Stelian Coros},
  \bibinfo{person}{Alexander Alspach}, \bibinfo{person}{James~M Bern},
  \bibinfo{person}{Joohyung Kim}, {and} \bibinfo{person}{Katsu Yamane}.}
  \bibinfo{year}{2018}\natexlab{}.
\newblock \showarticletitle{Computational design of robotic devices from
  high-level motion specifications}.
\newblock \bibinfo{journal}{\emph{IEEE Transactions on Robotics}}
  \bibinfo{volume}{34}, \bibinfo{number}{5} (\bibinfo{year}{2018}),
  \bibinfo{pages}{1240--1251}.
\newblock


\bibitem[\protect\citeauthoryear{Ha, Coros, Alspach, Kim, and Yamane}{Ha
  et~al\mbox{.}}{2017}]%
        {ha2017joint}
\bibfield{author}{\bibinfo{person}{Sehoon Ha}, \bibinfo{person}{Stelian Coros},
  \bibinfo{person}{Alexander Alspach}, \bibinfo{person}{Joohyung Kim}, {and}
  \bibinfo{person}{Katsu Yamane}.} \bibinfo{year}{2017}\natexlab{}.
\newblock \showarticletitle{Joint Optimization of Robot Design and Motion
  Parameters using the Implicit Function Theorem.}. In
  \bibinfo{booktitle}{\emph{Robotics: Science and systems}},
  Vol.~\bibinfo{volume}{8}.
\newblock


\bibitem[\protect\citeauthoryear{Hahn, Banzet, Bern, and Coros}{Hahn
  et~al\mbox{.}}{2019}]%
        {hahn2019real2sim}
\bibfield{author}{\bibinfo{person}{David Hahn}, \bibinfo{person}{Pol Banzet},
  \bibinfo{person}{James~M Bern}, {and} \bibinfo{person}{Stelian Coros}.}
  \bibinfo{year}{2019}\natexlab{}.
\newblock \showarticletitle{Real2sim: Visco-elastic parameter estimation from
  dynamic motion}.
\newblock \bibinfo{journal}{\emph{ACM Transactions on Graphics (TOG)}}
  \bibinfo{volume}{38}, \bibinfo{number}{6} (\bibinfo{year}{2019}),
  \bibinfo{pages}{1--13}.
\newblock


\bibitem[\protect\citeauthoryear{Hansen, M{\"u}ller, and Koumoutsakos}{Hansen
  et~al\mbox{.}}{2003}]%
        {hansen2003reducing}
\bibfield{author}{\bibinfo{person}{Nikolaus Hansen}, \bibinfo{person}{Sibylle~D
  M{\"u}ller}, {and} \bibinfo{person}{Petros Koumoutsakos}.}
  \bibinfo{year}{2003}\natexlab{}.
\newblock \showarticletitle{Reducing the time complexity of the derandomized
  evolution strategy with covariance matrix adaptation (CMA-ES)}.
\newblock \bibinfo{journal}{\emph{Evolutionary computation}}
  \bibinfo{volume}{11}, \bibinfo{number}{1} (\bibinfo{year}{2003}),
  \bibinfo{pages}{1--18}.
\newblock


\bibitem[\protect\citeauthoryear{Hecker, Raabe, Enslow, DeWeese, Maynard, and
  van Prooijen}{Hecker et~al\mbox{.}}{2008}]%
        {hecker2008real}
\bibfield{author}{\bibinfo{person}{Chris Hecker}, \bibinfo{person}{Bernd
  Raabe}, \bibinfo{person}{Ryan~W Enslow}, \bibinfo{person}{John DeWeese},
  \bibinfo{person}{Jordan Maynard}, {and} \bibinfo{person}{Kees van Prooijen}.}
  \bibinfo{year}{2008}\natexlab{}.
\newblock \showarticletitle{Real-time motion retargeting to highly varied
  user-created morphologies}.
\newblock \bibinfo{journal}{\emph{ACM Transactions on Graphics (TOG)}}
  \bibinfo{volume}{27}, \bibinfo{number}{3} (\bibinfo{year}{2008}),
  \bibinfo{pages}{1--11}.
\newblock


\bibitem[\protect\citeauthoryear{Holl, Thuerey, and Koltun}{Holl
  et~al\mbox{.}}{2020}]%
        {holl2020learning}
\bibfield{author}{\bibinfo{person}{Philipp Holl}, \bibinfo{person}{Nils
  Thuerey}, {and} \bibinfo{person}{Vladlen Koltun}.}
  \bibinfo{year}{2020}\natexlab{}.
\newblock \showarticletitle{Learning to Control PDEs with Differentiable
  Physics}. In \bibinfo{booktitle}{\emph{International Conference on Learning
  Representations}}.
\newblock


\bibitem[\protect\citeauthoryear{Hu, Anderson, Li, Sun, Carr, Ragan-Kelley, and
  Durand}{Hu et~al\mbox{.}}{2020}]%
        {hu2019difftaichi}
\bibfield{author}{\bibinfo{person}{Yuanming Hu}, \bibinfo{person}{Luke
  Anderson}, \bibinfo{person}{Tzu-Mao Li}, \bibinfo{person}{Qi Sun},
  \bibinfo{person}{Nathan Carr}, \bibinfo{person}{Jonathan Ragan-Kelley}, {and}
  \bibinfo{person}{Fr{\'e}do Durand}.} \bibinfo{year}{2020}\natexlab{}.
\newblock \showarticletitle{Diff{T}aichi: Differentiable Programming for
  Physical Simulation}.
\newblock \bibinfo{journal}{\emph{International Conference on Learning
  Representations}} (\bibinfo{year}{2020}).
\newblock


\bibitem[\protect\citeauthoryear{Hu, Liu, Spielberg, Tenenbaum, Freeman, Wu,
  Rus, and Matusik}{Hu et~al\mbox{.}}{2019}]%
        {hu2019chainqueen}
\bibfield{author}{\bibinfo{person}{Yuanming Hu}, \bibinfo{person}{Jiancheng
  Liu}, \bibinfo{person}{Andrew Spielberg}, \bibinfo{person}{Joshua~B
  Tenenbaum}, \bibinfo{person}{William~T Freeman}, \bibinfo{person}{Jiajun Wu},
  \bibinfo{person}{Daniela Rus}, {and} \bibinfo{person}{Wojciech Matusik}.}
  \bibinfo{year}{2019}\natexlab{}.
\newblock \showarticletitle{Chain{Q}ueen: A real-time differentiable physical
  simulator for soft robotics}. In \bibinfo{booktitle}{\emph{2019 International
  conference on robotics and automation (ICRA)}}. IEEE,
  \bibinfo{pages}{6265--6271}.
\newblock


\bibitem[\protect\citeauthoryear{Huang, Hu, Du, Zhou, Su, Tenenbaum, and
  Gan}{Huang et~al\mbox{.}}{2021}]%
        {huang2021plasticinelab}
\bibfield{author}{\bibinfo{person}{Zhiao Huang}, \bibinfo{person}{Yuanming Hu},
  \bibinfo{person}{Tao Du}, \bibinfo{person}{Siyuan Zhou}, \bibinfo{person}{Hao
  Su}, \bibinfo{person}{Joshua~B. Tenenbaum}, {and} \bibinfo{person}{Chuang
  Gan}.} \bibinfo{year}{2021}\natexlab{}.
\newblock \showarticletitle{PlasticineLab: A Soft-Body Manipulation Benchmark
  with Differentiable Physics}. In \bibinfo{booktitle}{\emph{International
  Conference on Learning Representations}}.
\newblock


\bibitem[\protect\citeauthoryear{Katzschmann, Della~Santina, Toshimitsu,
  Bicchi, and Rus}{Katzschmann et~al\mbox{.}}{2019a}]%
        {katzschmann2019dynamic}
\bibfield{author}{\bibinfo{person}{Robert~K Katzschmann},
  \bibinfo{person}{Cosimo Della~Santina}, \bibinfo{person}{Yasunori
  Toshimitsu}, \bibinfo{person}{Antonio Bicchi}, {and} \bibinfo{person}{Daniela
  Rus}.} \bibinfo{year}{2019}\natexlab{a}.
\newblock \showarticletitle{Dynamic motion control of multi-segment soft robots
  using piecewise constant curvature matched with an augmented rigid body
  model}. In \bibinfo{booktitle}{\emph{2019 2nd IEEE International Conference
  on Soft Robotics (RoboSoft)}}. IEEE, \bibinfo{pages}{454--461}.
\newblock


\bibitem[\protect\citeauthoryear{Katzschmann, DelPreto, MacCurdy, and
  Rus}{Katzschmann et~al\mbox{.}}{2018}]%
        {katzschmann2018exploration}
\bibfield{author}{\bibinfo{person}{Robert~K Katzschmann},
  \bibinfo{person}{Joseph DelPreto}, \bibinfo{person}{Robert MacCurdy}, {and}
  \bibinfo{person}{Daniela Rus}.} \bibinfo{year}{2018}\natexlab{}.
\newblock \showarticletitle{Exploration of underwater life with an acoustically
  controlled soft robotic fish}.
\newblock \bibinfo{journal}{\emph{Science Robotics}} \bibinfo{volume}{3},
  \bibinfo{number}{16} (\bibinfo{year}{2018}).
\newblock


\bibitem[\protect\citeauthoryear{Katzschmann, Thieffry, Goury, Kruszewski,
  Guerra, Duriez, and Rus}{Katzschmann et~al\mbox{.}}{2019b}]%
        {katzschmann2019dynamically}
\bibfield{author}{\bibinfo{person}{Robert~K Katzschmann},
  \bibinfo{person}{Maxime Thieffry}, \bibinfo{person}{Olivier Goury},
  \bibinfo{person}{Alexandre Kruszewski}, \bibinfo{person}{Thierry-Marie
  Guerra}, \bibinfo{person}{Christian Duriez}, {and} \bibinfo{person}{Daniela
  Rus}.} \bibinfo{year}{2019}\natexlab{b}.
\newblock \showarticletitle{Dynamically closed-loop controlled soft robotic arm
  using a reduced order finite element model with state observer}. In
  \bibinfo{booktitle}{\emph{2019 2nd IEEE International Conference on Soft
  Robotics (RoboSoft)}}. IEEE, \bibinfo{pages}{717--724}.
\newblock


\bibitem[\protect\citeauthoryear{Kingma and Ba}{Kingma and Ba}{2015}]%
        {kingma2014adam}
\bibfield{author}{\bibinfo{person}{Diederik~P Kingma} {and}
  \bibinfo{person}{Jimmy Ba}.} \bibinfo{year}{2015}\natexlab{}.
\newblock \showarticletitle{Adam: A Method for Stochastic Optimization}. In
  \bibinfo{booktitle}{\emph{International Conference on Learning
  Representations}}.
\newblock


\bibitem[\protect\citeauthoryear{Lewis, Anjyo, Rhee, Zhang, Pighin, and
  Deng}{Lewis et~al\mbox{.}}{2014}]%
        {lewis2014practice}
\bibfield{author}{\bibinfo{person}{John~P Lewis}, \bibinfo{person}{Ken Anjyo},
  \bibinfo{person}{Taehyun Rhee}, \bibinfo{person}{Mengjie Zhang},
  \bibinfo{person}{Frederic~H Pighin}, {and} \bibinfo{person}{Zhigang Deng}.}
  \bibinfo{year}{2014}\natexlab{}.
\newblock \showarticletitle{Practice and Theory of Blendshape Facial Models.}
\newblock \bibinfo{journal}{\emph{Eurographics (State of the Art Reports)}}
  \bibinfo{volume}{1}, \bibinfo{number}{8} (\bibinfo{year}{2014}),
  \bibinfo{pages}{2}.
\newblock


\bibitem[\protect\citeauthoryear{Li, Wu, Tedrake, Tenenbaum, and Torralba}{Li
  et~al\mbox{.}}{2019}]%
        {li2018learning}
\bibfield{author}{\bibinfo{person}{Yunzhu Li}, \bibinfo{person}{Jiajun Wu},
  \bibinfo{person}{Russ Tedrake}, \bibinfo{person}{Joshua Tenenbaum}, {and}
  \bibinfo{person}{Antonio Torralba}.} \bibinfo{year}{2019}\natexlab{}.
\newblock \showarticletitle{Learning Particle Dynamics for Manipulating Rigid
  Bodies, Deformable Objects, and Fluids}. In
  \bibinfo{booktitle}{\emph{International Conference on Learning
  Representations}}.
\newblock


\bibitem[\protect\citeauthoryear{Liang, Lin, and Koltun}{Liang
  et~al\mbox{.}}{2019}]%
        {liang2019differentiable}
\bibfield{author}{\bibinfo{person}{Junbang Liang}, \bibinfo{person}{Ming Lin},
  {and} \bibinfo{person}{Vladlen Koltun}.} \bibinfo{year}{2019}\natexlab{}.
\newblock \showarticletitle{Differentiable Cloth Simulation for Inverse
  Problems}. In \bibinfo{booktitle}{\emph{Advances in Neural Information
  Processing Systems}}. \bibinfo{pages}{772--781}.
\newblock


\bibitem[\protect\citeauthoryear{Lighthill}{Lighthill}{1971}]%
        {lighthill1971large}
\bibfield{author}{\bibinfo{person}{Michael~James Lighthill}.}
  \bibinfo{year}{1971}\natexlab{}.
\newblock \showarticletitle{Large-amplitude elongated-body theory of fish
  locomotion}.
\newblock \bibinfo{journal}{\emph{Proceedings of the Royal Society of London.
  Series B. Biological Sciences}} \bibinfo{volume}{179}, \bibinfo{number}{1055}
  (\bibinfo{year}{1971}), \bibinfo{pages}{125--138}.
\newblock


\bibitem[\protect\citeauthoryear{Ma, Tian, Pan, Ren, and Manocha}{Ma
  et~al\mbox{.}}{2018}]%
        {ma2018fluid}
\bibfield{author}{\bibinfo{person}{Pingchuan Ma}, \bibinfo{person}{Yunsheng
  Tian}, \bibinfo{person}{Zherong Pan}, \bibinfo{person}{Bo Ren}, {and}
  \bibinfo{person}{Dinesh Manocha}.} \bibinfo{year}{2018}\natexlab{}.
\newblock \showarticletitle{Fluid directed rigid body control using deep
  reinforcement learning}.
\newblock \bibinfo{journal}{\emph{ACM Transactions on Graphics (TOG)}}
  \bibinfo{volume}{37}, \bibinfo{number}{4} (\bibinfo{year}{2018}),
  \bibinfo{pages}{1--11}.
\newblock


\bibitem[\protect\citeauthoryear{Marchese, Onal, and Rus}{Marchese
  et~al\mbox{.}}{2014}]%
        {marchese2014autonomous}
\bibfield{author}{\bibinfo{person}{Andrew~D Marchese},
  \bibinfo{person}{Cagdas~D Onal}, {and} \bibinfo{person}{Daniela Rus}.}
  \bibinfo{year}{2014}\natexlab{}.
\newblock \showarticletitle{Autonomous soft robotic fish capable of escape
  maneuvers using fluidic elastomer actuators}.
\newblock \bibinfo{journal}{\emph{Soft robotics}} \bibinfo{volume}{1},
  \bibinfo{number}{1} (\bibinfo{year}{2014}), \bibinfo{pages}{75--87}.
\newblock


\bibitem[\protect\citeauthoryear{Marchese, Tedrake, and Rus}{Marchese
  et~al\mbox{.}}{2016}]%
        {marchese2016dynamics}
\bibfield{author}{\bibinfo{person}{Andrew~D Marchese}, \bibinfo{person}{Russ
  Tedrake}, {and} \bibinfo{person}{Daniela Rus}.}
  \bibinfo{year}{2016}\natexlab{}.
\newblock \showarticletitle{Dynamics and trajectory optimization for a soft
  spatial fluidic elastomer manipulator}.
\newblock \bibinfo{journal}{\emph{The International Journal of Robotics
  Research}} \bibinfo{volume}{35}, \bibinfo{number}{8} (\bibinfo{year}{2016}),
  \bibinfo{pages}{1000--1019}.
\newblock


\bibitem[\protect\citeauthoryear{McNamara, Treuille, Popovi{\'c}, and
  Stam}{McNamara et~al\mbox{.}}{2004}]%
        {mcnamara2004fluid}
\bibfield{author}{\bibinfo{person}{Antoine McNamara}, \bibinfo{person}{Adrien
  Treuille}, \bibinfo{person}{Zoran Popovi{\'c}}, {and} \bibinfo{person}{Jos
  Stam}.} \bibinfo{year}{2004}\natexlab{}.
\newblock \showarticletitle{Fluid control using the adjoint method}.
\newblock \bibinfo{journal}{\emph{ACM Transactions On Graphics (TOG)}}
  \bibinfo{volume}{23}, \bibinfo{number}{3} (\bibinfo{year}{2004}),
  \bibinfo{pages}{449--456}.
\newblock


\bibitem[\protect\citeauthoryear{Megaro, Thomaszewski, Nitti, Hilliges, Gross,
  and Coros}{Megaro et~al\mbox{.}}{2015}]%
        {megaro2015interactive}
\bibfield{author}{\bibinfo{person}{Vittorio Megaro}, \bibinfo{person}{Bernhard
  Thomaszewski}, \bibinfo{person}{Maurizio Nitti}, \bibinfo{person}{Otmar
  Hilliges}, \bibinfo{person}{Markus Gross}, {and} \bibinfo{person}{Stelian
  Coros}.} \bibinfo{year}{2015}\natexlab{}.
\newblock \showarticletitle{Interactive design of 3d-printable robotic
  creatures}.
\newblock \bibinfo{journal}{\emph{ACM Transactions on Graphics (TOG)}}
  \bibinfo{volume}{34}, \bibinfo{number}{6} (\bibinfo{year}{2015}),
  \bibinfo{pages}{1--9}.
\newblock


\bibitem[\protect\citeauthoryear{Min, Won, Lee, Park, and Lee}{Min
  et~al\mbox{.}}{2019}]%
        {min2019softcon}
\bibfield{author}{\bibinfo{person}{Sehee Min}, \bibinfo{person}{Jungdam Won},
  \bibinfo{person}{Seunghwan Lee}, \bibinfo{person}{Jungnam Park}, {and}
  \bibinfo{person}{Jehee Lee}.} \bibinfo{year}{2019}\natexlab{}.
\newblock \showarticletitle{SoftCon: simulation and control of soft-bodied
  animals with biomimetic actuators}.
\newblock \bibinfo{journal}{\emph{ACM Transactions on Graphics (TOG)}}
  \bibinfo{volume}{38}, \bibinfo{number}{6} (\bibinfo{year}{2019}),
  \bibinfo{pages}{1--12}.
\newblock


\bibitem[\protect\citeauthoryear{Mo, Guerrero, Yi, Su, Wonka, Mitra, and
  Guibas}{Mo et~al\mbox{.}}{2019}]%
        {mo2019structurenet}
\bibfield{author}{\bibinfo{person}{Kaichun Mo}, \bibinfo{person}{Paul
  Guerrero}, \bibinfo{person}{Li Yi}, \bibinfo{person}{Hao Su},
  \bibinfo{person}{Peter Wonka}, \bibinfo{person}{Niloy~J Mitra}, {and}
  \bibinfo{person}{Leonidas~J Guibas}.} \bibinfo{year}{2019}\natexlab{}.
\newblock \showarticletitle{StructureNet: hierarchical graph networks for 3D
  shape generation}.
\newblock \bibinfo{journal}{\emph{ACM Transactions on Graphics (TOG)}}
  \bibinfo{volume}{38}, \bibinfo{number}{6} (\bibinfo{year}{2019}),
  \bibinfo{pages}{1--19}.
\newblock


\bibitem[\protect\citeauthoryear{Ovsjanikov, Ben-Chen, Solomon, Butscher, and
  Guibas}{Ovsjanikov et~al\mbox{.}}{2012}]%
        {ovsjanikov2012functional}
\bibfield{author}{\bibinfo{person}{Maks Ovsjanikov}, \bibinfo{person}{Mirela
  Ben-Chen}, \bibinfo{person}{Justin Solomon}, \bibinfo{person}{Adrian
  Butscher}, {and} \bibinfo{person}{Leonidas Guibas}.}
  \bibinfo{year}{2012}\natexlab{}.
\newblock \showarticletitle{Functional maps: a flexible representation of maps
  between shapes}.
\newblock \bibinfo{journal}{\emph{ACM Transactions on Graphics (TOG)}}
  \bibinfo{volume}{31}, \bibinfo{number}{4} (\bibinfo{year}{2012}),
  \bibinfo{pages}{1--11}.
\newblock


\bibitem[\protect\citeauthoryear{Park, Florence, Straub, Newcombe, and
  Lovegrove}{Park et~al\mbox{.}}{2019}]%
        {park2019deepsdf}
\bibfield{author}{\bibinfo{person}{Jeong~Joon Park}, \bibinfo{person}{Peter
  Florence}, \bibinfo{person}{Julian Straub}, \bibinfo{person}{Richard
  Newcombe}, {and} \bibinfo{person}{Steven Lovegrove}.}
  \bibinfo{year}{2019}\natexlab{}.
\newblock \showarticletitle{Deepsdf: Learning continuous signed distance
  functions for shape representation}. In \bibinfo{booktitle}{\emph{Proceedings
  of the IEEE/CVF Conference on Computer Vision and Pattern Recognition}}.
  \bibinfo{pages}{165--174}.
\newblock


\bibitem[\protect\citeauthoryear{Paszke, Gross, Massa, Lerer, Bradbury, Chanan,
  Killeen, Lin, Gimelshein, Antiga, et~al\mbox{.}}{Paszke
  et~al\mbox{.}}{2019}]%
        {paszke2019pytorch}
\bibfield{author}{\bibinfo{person}{Adam Paszke}, \bibinfo{person}{Sam Gross},
  \bibinfo{person}{Francisco Massa}, \bibinfo{person}{Adam Lerer},
  \bibinfo{person}{James Bradbury}, \bibinfo{person}{Gregory Chanan},
  \bibinfo{person}{Trevor Killeen}, \bibinfo{person}{Zeming Lin},
  \bibinfo{person}{Natalia Gimelshein}, \bibinfo{person}{Luca Antiga},
  {et~al\mbox{.}}} \bibinfo{year}{2019}\natexlab{}.
\newblock \showarticletitle{PyTorch: An Imperative Style, High-Performance Deep
  Learning Library}.
\newblock \bibinfo{journal}{\emph{Advances in Neural Information Processing
  Systems}}  \bibinfo{volume}{32} (\bibinfo{year}{2019}),
  \bibinfo{pages}{8026--8037}.
\newblock


\bibitem[\protect\citeauthoryear{Pathak, Lu, Darrell, Isola, and Efros}{Pathak
  et~al\mbox{.}}{2019}]%
        {pathak2019learning}
\bibfield{author}{\bibinfo{person}{Deepak Pathak}, \bibinfo{person}{Christopher
  Lu}, \bibinfo{person}{Trevor Darrell}, \bibinfo{person}{Phillip Isola}, {and}
  \bibinfo{person}{Alexei~A Efros}.} \bibinfo{year}{2019}\natexlab{}.
\newblock \showarticletitle{Learning to Control Self-Assembling Morphologies: A
  Study of Generalization via Modularity}.
\newblock \bibinfo{journal}{\emph{Advances in Neural Information Processing
  Systems}}  \bibinfo{volume}{32} (\bibinfo{year}{2019}),
  \bibinfo{pages}{2295--2305}.
\newblock


\bibitem[\protect\citeauthoryear{Popovi{\'c}, Seitz, and Erdmann}{Popovi{\'c}
  et~al\mbox{.}}{2003}]%
        {popovic2003motion}
\bibfield{author}{\bibinfo{person}{Jovan Popovi{\'c}},
  \bibinfo{person}{Steven~M Seitz}, {and} \bibinfo{person}{Michael Erdmann}.}
  \bibinfo{year}{2003}\natexlab{}.
\newblock \showarticletitle{Motion sketching for control of rigid-body
  simulations}.
\newblock \bibinfo{journal}{\emph{ACM Transactions on Graphics (TOG)}}
  \bibinfo{volume}{22}, \bibinfo{number}{4} (\bibinfo{year}{2003}),
  \bibinfo{pages}{1034--1054}.
\newblock


\bibitem[\protect\citeauthoryear{Qiao, Liang, Koltun, and Lin}{Qiao
  et~al\mbox{.}}{2020}]%
        {qiao2020scalable}
\bibfield{author}{\bibinfo{person}{Yi-Ling Qiao}, \bibinfo{person}{Junbang
  Liang}, \bibinfo{person}{Vladlen Koltun}, {and} \bibinfo{person}{Ming Lin}.}
  \bibinfo{year}{2020}\natexlab{}.
\newblock \showarticletitle{Scalable Differentiable Physics for Learning and
  Control}. In \bibinfo{booktitle}{\emph{International Conference on Machine
  Learning}}.
\newblock


\bibitem[\protect\citeauthoryear{Rubner, Tomasi, and Guibas}{Rubner
  et~al\mbox{.}}{2000}]%
        {rubner2000earth}
\bibfield{author}{\bibinfo{person}{Yossi Rubner}, \bibinfo{person}{Carlo
  Tomasi}, {and} \bibinfo{person}{Leonidas~J Guibas}.}
  \bibinfo{year}{2000}\natexlab{}.
\newblock \showarticletitle{The earth mover's distance as a metric for image
  retrieval}.
\newblock \bibinfo{journal}{\emph{International journal of computer vision}}
  \bibinfo{volume}{40}, \bibinfo{number}{2} (\bibinfo{year}{2000}),
  \bibinfo{pages}{99--121}.
\newblock


\bibitem[\protect\citeauthoryear{Sanchez-Gonzalez, Godwin, Pfaff, Ying,
  Leskovec, and Battaglia}{Sanchez-Gonzalez et~al\mbox{.}}{2020}]%
        {sanchez2020learning}
\bibfield{author}{\bibinfo{person}{Alvaro Sanchez-Gonzalez},
  \bibinfo{person}{Jonathan Godwin}, \bibinfo{person}{Tobias Pfaff},
  \bibinfo{person}{Rex Ying}, \bibinfo{person}{Jure Leskovec}, {and}
  \bibinfo{person}{Peter Battaglia}.} \bibinfo{year}{2020}\natexlab{}.
\newblock \showarticletitle{Learning to Simulate Complex Physics with Graph
  Networks}. In \bibinfo{booktitle}{\emph{International Conference on Machine
  Learning}}.
\newblock


\bibitem[\protect\citeauthoryear{Schaff, Yunis, Chakrabarti, and Walter}{Schaff
  et~al\mbox{.}}{2019}]%
        {schaff2019jointly}
\bibfield{author}{\bibinfo{person}{Charles Schaff}, \bibinfo{person}{David
  Yunis}, \bibinfo{person}{Ayan Chakrabarti}, {and} \bibinfo{person}{Matthew~R
  Walter}.} \bibinfo{year}{2019}\natexlab{}.
\newblock \showarticletitle{Jointly learning to construct and control agents
  using deep reinforcement learning}. In \bibinfo{booktitle}{\emph{2019
  International Conference on Robotics and Automation (ICRA)}}. IEEE,
  \bibinfo{pages}{9798--9805}.
\newblock


\bibitem[\protect\citeauthoryear{Schenck and Fox}{Schenck and Fox}{2018}]%
        {schenck2018spnets}
\bibfield{author}{\bibinfo{person}{Connor Schenck} {and}
  \bibinfo{person}{Dieter Fox}.} \bibinfo{year}{2018}\natexlab{}.
\newblock \showarticletitle{{SPN}ets: Differentiable Fluid Dynamics for Deep
  Neural Networks}.
\newblock \bibinfo{journal}{\emph{Conference on Robot Learning (CoRL)}}
  (\bibinfo{year}{2018}).
\newblock


\bibitem[\protect\citeauthoryear{Schlick}{Schlick}{1994}]%
        {schlick1994fast}
\bibfield{author}{\bibinfo{person}{Christophe Schlick}.}
  \bibinfo{year}{1994}\natexlab{}.
\newblock \showarticletitle{Fast Alternatives to Perlin's Bias and Gain
  Functions.}
\newblock \bibinfo{journal}{\emph{Graphics Gems}}  \bibinfo{volume}{4}
  (\bibinfo{year}{1994}), \bibinfo{pages}{401--404}.
\newblock


\bibitem[\protect\citeauthoryear{Schulz, Shamir, Baran, Levin, Sitthi-Amorn,
  and Matusik}{Schulz et~al\mbox{.}}{2017a}]%
        {schulz2017retrieval}
\bibfield{author}{\bibinfo{person}{Adriana Schulz}, \bibinfo{person}{Ariel
  Shamir}, \bibinfo{person}{Ilya Baran}, \bibinfo{person}{David~IW Levin},
  \bibinfo{person}{Pitchaya Sitthi-Amorn}, {and} \bibinfo{person}{Wojciech
  Matusik}.} \bibinfo{year}{2017}\natexlab{a}.
\newblock \showarticletitle{Retrieval on parametric shape collections}.
\newblock \bibinfo{journal}{\emph{ACM Transactions on Graphics (TOG)}}
  \bibinfo{volume}{36}, \bibinfo{number}{1} (\bibinfo{year}{2017}),
  \bibinfo{pages}{1--14}.
\newblock


\bibitem[\protect\citeauthoryear{Schulz, Sung, Spielberg, Zhao, Cheng,
  Grinspun, Rus, and Matusik}{Schulz et~al\mbox{.}}{2017b}]%
        {schulz2017interactive}
\bibfield{author}{\bibinfo{person}{Adriana Schulz}, \bibinfo{person}{Cynthia
  Sung}, \bibinfo{person}{Andrew Spielberg}, \bibinfo{person}{Wei Zhao},
  \bibinfo{person}{Robin Cheng}, \bibinfo{person}{Eitan Grinspun},
  \bibinfo{person}{Daniela Rus}, {and} \bibinfo{person}{Wojciech Matusik}.}
  \bibinfo{year}{2017}\natexlab{b}.
\newblock \showarticletitle{Interactive robogami: An end-to-end system for
  design of robots with ground locomotion}.
\newblock \bibinfo{journal}{\emph{The International Journal of Robotics
  Research}} \bibinfo{volume}{36}, \bibinfo{number}{10} (\bibinfo{year}{2017}),
  \bibinfo{pages}{1131--1147}.
\newblock


\bibitem[\protect\citeauthoryear{Sfakiotakis, Lane, and Davies}{Sfakiotakis
  et~al\mbox{.}}{1999}]%
        {sfakiotakis1999review}
\bibfield{author}{\bibinfo{person}{Michael Sfakiotakis},
  \bibinfo{person}{David~M Lane}, {and} \bibinfo{person}{J~Bruce~C Davies}.}
  \bibinfo{year}{1999}\natexlab{}.
\newblock \showarticletitle{Review of fish swimming modes for aquatic
  locomotion}.
\newblock \bibinfo{journal}{\emph{IEEE Journal of oceanic engineering}}
  \bibinfo{volume}{24}, \bibinfo{number}{2} (\bibinfo{year}{1999}),
  \bibinfo{pages}{237--252}.
\newblock


\bibitem[\protect\citeauthoryear{Sims}{Sims}{1994}]%
        {sims1994evolving}
\bibfield{author}{\bibinfo{person}{Karl Sims}.}
  \bibinfo{year}{1994}\natexlab{}.
\newblock \showarticletitle{Evolving virtual creatures}. In
  \bibinfo{booktitle}{\emph{Proceedings of the 21st annual conference on
  Computer graphics and interactive techniques}}. \bibinfo{pages}{15--22}.
\newblock


\bibitem[\protect\citeauthoryear{Solomon, De~Goes, Peyr{\'e}, Cuturi, Butscher,
  Nguyen, Du, and Guibas}{Solomon et~al\mbox{.}}{2015}]%
        {solomon2015}
\bibfield{author}{\bibinfo{person}{Justin Solomon}, \bibinfo{person}{Fernando
  De~Goes}, \bibinfo{person}{Gabriel Peyr{\'e}}, \bibinfo{person}{Marco
  Cuturi}, \bibinfo{person}{Adrian Butscher}, \bibinfo{person}{Andy Nguyen},
  \bibinfo{person}{Tao Du}, {and} \bibinfo{person}{Leonidas Guibas}.}
  \bibinfo{year}{2015}\natexlab{}.
\newblock \showarticletitle{Convolutional wasserstein distances: Efficient
  optimal transportation on geometric domains}.
\newblock \bibinfo{journal}{\emph{ACM Transactions on Graphics (TOG)}}
  \bibinfo{volume}{34}, \bibinfo{number}{4} (\bibinfo{year}{2015}),
  \bibinfo{pages}{1--11}.
\newblock


\bibitem[\protect\citeauthoryear{Spielberg, Araki, Sung, Tedrake, and
  Rus}{Spielberg et~al\mbox{.}}{2017}]%
        {spielberg2017functional}
\bibfield{author}{\bibinfo{person}{Andrew Spielberg}, \bibinfo{person}{Brandon
  Araki}, \bibinfo{person}{Cynthia Sung}, \bibinfo{person}{Russ Tedrake}, {and}
  \bibinfo{person}{Daniela Rus}.} \bibinfo{year}{2017}\natexlab{}.
\newblock \showarticletitle{Functional co-optimization of articulated robots}.
  In \bibinfo{booktitle}{\emph{2017 IEEE International Conference on Robotics
  and Automation (ICRA)}}. IEEE, \bibinfo{pages}{5035--5042}.
\newblock


\bibitem[\protect\citeauthoryear{Spielberg, Zhao, Hu, Du, Matusik, and
  Rus}{Spielberg et~al\mbox{.}}{2019}]%
        {spielberg2019learning}
\bibfield{author}{\bibinfo{person}{Andrew Spielberg}, \bibinfo{person}{Allan
  Zhao}, \bibinfo{person}{Yuanming Hu}, \bibinfo{person}{Tao Du},
  \bibinfo{person}{Wojciech Matusik}, {and} \bibinfo{person}{Daniela Rus}.}
  \bibinfo{year}{2019}\natexlab{}.
\newblock \showarticletitle{Learning-in-the-loop optimization: End-to-end
  control and co-design of soft robots through learned deep latent
  representations}.
\newblock \bibinfo{journal}{\emph{Advances in Neural Information Processing
  Systems}}  \bibinfo{volume}{32} (\bibinfo{year}{2019}),
  \bibinfo{pages}{8284--8294}.
\newblock


\bibitem[\protect\citeauthoryear{Tan, Gu, Turk, and Liu}{Tan
  et~al\mbox{.}}{2011}]%
        {tan2011articulated}
\bibfield{author}{\bibinfo{person}{Jie Tan}, \bibinfo{person}{Yuting Gu},
  \bibinfo{person}{Greg Turk}, {and} \bibinfo{person}{C~Karen Liu}.}
  \bibinfo{year}{2011}\natexlab{}.
\newblock \showarticletitle{Articulated swimming creatures}.
\newblock \bibinfo{journal}{\emph{ACM Transactions on Graphics (TOG)}}
  \bibinfo{volume}{30}, \bibinfo{number}{4} (\bibinfo{year}{2011}),
  \bibinfo{pages}{1--12}.
\newblock


\bibitem[\protect\citeauthoryear{Thieffry, Kruszewski, Duriez, and
  Guerra}{Thieffry et~al\mbox{.}}{2018}]%
        {thieffry2018control}
\bibfield{author}{\bibinfo{person}{Maxime Thieffry}, \bibinfo{person}{Alexandre
  Kruszewski}, \bibinfo{person}{Christian Duriez}, {and}
  \bibinfo{person}{Thierry-Marie Guerra}.} \bibinfo{year}{2018}\natexlab{}.
\newblock \showarticletitle{Control Design for Soft Robots Based on
  Reduced-Order Model}.
\newblock \bibinfo{journal}{\emph{IEEE Robotics and Automation Letters}}
  \bibinfo{volume}{4}, \bibinfo{number}{1} (\bibinfo{year}{2018}),
  \bibinfo{pages}{25--32}.
\newblock


\bibitem[\protect\citeauthoryear{Triantafyllou and Triantafyllou}{Triantafyllou
  and Triantafyllou}{1995}]%
        {triantafyllou1995efficient}
\bibfield{author}{\bibinfo{person}{Michael~S Triantafyllou} {and}
  \bibinfo{person}{George~S Triantafyllou}.} \bibinfo{year}{1995}\natexlab{}.
\newblock \showarticletitle{An efficient swimming machine}.
\newblock \bibinfo{journal}{\emph{Scientific american}} \bibinfo{volume}{272},
  \bibinfo{number}{3} (\bibinfo{year}{1995}), \bibinfo{pages}{64--70}.
\newblock


\bibitem[\protect\citeauthoryear{Van~Diepen and Shea}{Van~Diepen and
  Shea}{2019}]%
        {van2019spatial}
\bibfield{author}{\bibinfo{person}{Merel Van~Diepen} {and}
  \bibinfo{person}{Kristina Shea}.} \bibinfo{year}{2019}\natexlab{}.
\newblock \showarticletitle{A spatial grammar method for the computational
  design synthesis of virtual soft locomotion robots}.
\newblock \bibinfo{journal}{\emph{Journal of Mechanical Design}}
  \bibinfo{volume}{141}, \bibinfo{number}{10} (\bibinfo{year}{2019}).
\newblock


\bibitem[\protect\citeauthoryear{Vaswani, Shazeer, Parmar, Uszkoreit, Jones,
  Gomez, Kaiser, and Polosukhin}{Vaswani et~al\mbox{.}}{2017}]%
        {vaswani2017attention}
\bibfield{author}{\bibinfo{person}{Ashish Vaswani}, \bibinfo{person}{Noam
  Shazeer}, \bibinfo{person}{Niki Parmar}, \bibinfo{person}{Jakob Uszkoreit},
  \bibinfo{person}{Llion Jones}, \bibinfo{person}{Aidan~N Gomez},
  \bibinfo{person}{{\L}ukasz Kaiser}, {and} \bibinfo{person}{Illia
  Polosukhin}.} \bibinfo{year}{2017}\natexlab{}.
\newblock \showarticletitle{Attention is all you need}. In
  \bibinfo{booktitle}{\emph{Proceedings of the 31st International Conference on
  Neural Information Processing Systems}}. \bibinfo{pages}{6000--6010}.
\newblock


\bibitem[\protect\citeauthoryear{Wampler and Popovi{\'c}}{Wampler and
  Popovi{\'c}}{2009}]%
        {wampler2009optimal}
\bibfield{author}{\bibinfo{person}{Kevin Wampler} {and} \bibinfo{person}{Zoran
  Popovi{\'c}}.} \bibinfo{year}{2009}\natexlab{}.
\newblock \showarticletitle{Optimal gait and form for animal locomotion}.
\newblock \bibinfo{journal}{\emph{ACM Transactions on Graphics (TOG)}}
  \bibinfo{volume}{28}, \bibinfo{number}{3} (\bibinfo{year}{2009}),
  \bibinfo{pages}{1--8}.
\newblock


\bibitem[\protect\citeauthoryear{Wang, Zhou, Fidler, and Ba}{Wang
  et~al\mbox{.}}{2018}]%
        {wang2019neural}
\bibfield{author}{\bibinfo{person}{Tingwu Wang}, \bibinfo{person}{Yuhao Zhou},
  \bibinfo{person}{Sanja Fidler}, {and} \bibinfo{person}{Jimmy Ba}.}
  \bibinfo{year}{2018}\natexlab{}.
\newblock \showarticletitle{Neural Graph Evolution: Towards Efficient Automatic
  Robot Design}. In \bibinfo{booktitle}{\emph{International Conference on
  Learning Representations}}.
\newblock


\bibitem[\protect\citeauthoryear{Won and Lee}{Won and Lee}{2019}]%
        {won2019learning}
\bibfield{author}{\bibinfo{person}{Jungdam Won} {and} \bibinfo{person}{Jehee
  Lee}.} \bibinfo{year}{2019}\natexlab{}.
\newblock \showarticletitle{Learning body shape variation in physics-based
  characters}.
\newblock \bibinfo{journal}{\emph{ACM Transactions on Graphics (TOG)}}
  \bibinfo{volume}{38}, \bibinfo{number}{6} (\bibinfo{year}{2019}),
  \bibinfo{pages}{1--12}.
\newblock


\bibitem[\protect\citeauthoryear{Yang, Huang, Hao, Liu, Belongie, and
  Hariharan}{Yang et~al\mbox{.}}{2019}]%
        {yang2019pointflow}
\bibfield{author}{\bibinfo{person}{Guandao Yang}, \bibinfo{person}{Xun Huang},
  \bibinfo{person}{Zekun Hao}, \bibinfo{person}{Ming-Yu Liu},
  \bibinfo{person}{Serge Belongie}, {and} \bibinfo{person}{Bharath Hariharan}.}
  \bibinfo{year}{2019}\natexlab{}.
\newblock \showarticletitle{Pointflow: 3d point cloud generation with
  continuous normalizing flows}. In \bibinfo{booktitle}{\emph{Proceedings of
  the IEEE/CVF International Conference on Computer Vision}}.
  \bibinfo{pages}{4541--4550}.
\newblock


\bibitem[\protect\citeauthoryear{Zhao, Xu, Konakovi{\'c}-Lukovi{\'c}, Hughes,
  Spielberg, Rus, and Matusik}{Zhao et~al\mbox{.}}{2020}]%
        {zhao2020}
\bibfield{author}{\bibinfo{person}{Allan Zhao}, \bibinfo{person}{Jie Xu},
  \bibinfo{person}{Mina Konakovi{\'c}-Lukovi{\'c}}, \bibinfo{person}{Josephine
  Hughes}, \bibinfo{person}{Andrew Spielberg}, \bibinfo{person}{Daniela Rus},
  {and} \bibinfo{person}{Wojciech Matusik}.} \bibinfo{year}{2020}\natexlab{}.
\newblock \showarticletitle{RoboGrammar: graph grammar for terrain-optimized
  robot design}.
\newblock \bibinfo{journal}{\emph{ACM Transactions on Graphics (TOG)}}
  \bibinfo{volume}{39}, \bibinfo{number}{6} (\bibinfo{year}{2020}),
  \bibinfo{pages}{1--16}.
\newblock


\end{thebibliography}

\end{document}